\RequirePackage{fix-cm}
\documentclass[twocolumn]{svjour3}          

\smartqed  
\usepackage{times}
\usepackage{epsfig}
\usepackage{graphicx}%
\usepackage{multirow}%
\usepackage[ruled,vlined]{algorithm2e}
\usepackage{amsmath,amssymb,amsfonts}%
\usepackage{bm}
\usepackage{textcomp}
\usepackage{enumitem}
\usepackage[misc]{ifsym} 

\usepackage[pagebackref=true,breaklinks=true,letterpaper=true,colorlinks,bookmarks=false]{hyperref}
\usepackage[numbers]{natbib}

\usepackage{caption}
\usepackage{booktabs}
\usepackage{comment}
\usepackage{multirow}
\usepackage{color,xcolor}
\usepackage{wrapfig}
\usepackage{footmisc}

\definecolor{mygreen}{rgb}{0.09, 0.45, 0.27}

\usepackage{multirow}
\usepackage[pagebackref=true,breaklinks=true,colorlinks,bookmarks=false]{hyperref}
\usepackage{arydshln} 
\usepackage{orcidlink}
\usepackage{pdfrender}

\usepackage{wrapfig}
\usepackage{subfig}
\usepackage{tablefootnote}
\usepackage{bm}

\usepackage{colortbl} 
\definecolor{mygray}{gray}{0.6}
\definecolor{mygray-bg}{gray}{0.9}
\definecolor{mygray-u}{gray}{0.92}

\newcommand{\ie}{\textit{i}.\textit{e}.}
\newcommand{\eg}{\textit{e}.\textit{g}.}
\newcommand{\cf}{\textit{cf.}}

\newcommand{\vs}{\textit{vs}.}

\definecolor{mygray}{gray}{0.6}
\definecolor{mygray-bg}{gray}{0.9}
\usepackage{bbding}
\usepackage{pifont}
\usepackage{wasysym}
\usepackage{booktabs}

\newcommand{\modified}[1]{{\color{black}{#1}}}

\begin{document}
\begin{sloppypar}
\title{From Easy to Hard: Learning Curricular Shape-aware Features for Robust Panoptic Scene Graph Generation\thanks{This work was supported by the National Key Research \& Development Project of China (2021ZD0110700), the National Natural Science Foundation of China (62337001) and the Fundamental Research Funds for the Central Universities (No. 226-2022-00051). Long Chen was supported by HKUST Special Support for Young Faculty (F0927) and HKUST Sports Science and Technology Research Grant (SSTRG24EG04).}}


\author{Hanrong Shi*, Lin Li$^\textrm{\Letter}$*, Jun Xiao, Yueting Zhuang and Long Chen$^\textrm{\Letter}$}


\institute{Hanrong Shi \at
              College of Computer Science \\ 
              Zhejiang University, Hangzhou, China \\
              \email{hanrong@zju.edu.cn}           
          \and
          Lin Li \at
              College of Computer Science \\
              Zhejiang University, Hangzhou, China \\
              \email{mukti@zju.edu.cn}
          \and
          Jun Xiao \at
              College of Computer Science \\
              Zhejiang University, Hangzhou, China \\
              \email{junx@cs.zju.edu.cn}
          \and
          Yueting Zhuang \at
              College of Computer Science \\
              Zhejiang University, Hangzhou, China \\
              \email{yzhuang@zju.edu.cn}
          \and
          Long Chen \at
              Department of Computer Science and Engineering \\
              The Hong Kong University of Science and Technology, Hong Kong \\
              \email{longchen@ust.hk} \\
        $^\textrm{\Letter}$Long Chen and Lin Li are the corresponding authors.\\
        * These authors contributed equally to this work.
}

\date{Received: date / Accepted: date}

\maketitle
\begin{abstract}
Panoptic Scene Graph Generation (PSG) aims to generate a comprehensive graph-structure representation based on panoptic segmentation masks. Despite remarkable progress in PSG, almost all existing methods neglect the importance of shape-aware features, which inherently focus on the contours and boundaries of objects. 
To bridge this gap, we propose a model-agnostic Curricular shApe-aware FEature (\textbf{CAFE}) learning strategy for PSG. Specifically, we incorporate shape-aware features (\ie, mask features and boundary features) into PSG, moving beyond reliance solely on bbox features. 
Furthermore, drawing inspiration from human cognition, we propose to integrate shape-aware features in an easy-to-hard manner. To achieve this, we categorize the predicates into three groups based on cognition learning difficulty and correspondingly divide the training process into three stages. Each stage utilizes a specialized relation classifier to distinguish specific groups of predicates. As the learning difficulty of predicates increases, these classifiers are equipped with features of ascending complexity. We also incorporate knowledge distillation to retain knowledge acquired in earlier stages. 
Due to its model-agnostic nature, CAFE can be seamlessly incorporated into any PSG model. Extensive experiments and ablations on two PSG tasks under both robust and zero-shot PSG have attested to the superiority and robustness of our proposed CAFE, which outperforms existing state-of-the-art methods by a large margin.
\keywords{Panoptic Scene Graph Generation \and Robust Learning \and Shape-aware Features \and Curriculum Learning \and Novel Class Discovery}
\end{abstract}

\section{Introduction}
\label{sec1}
\begin{figure}[t]
  \centering
  \includegraphics[width=\linewidth]{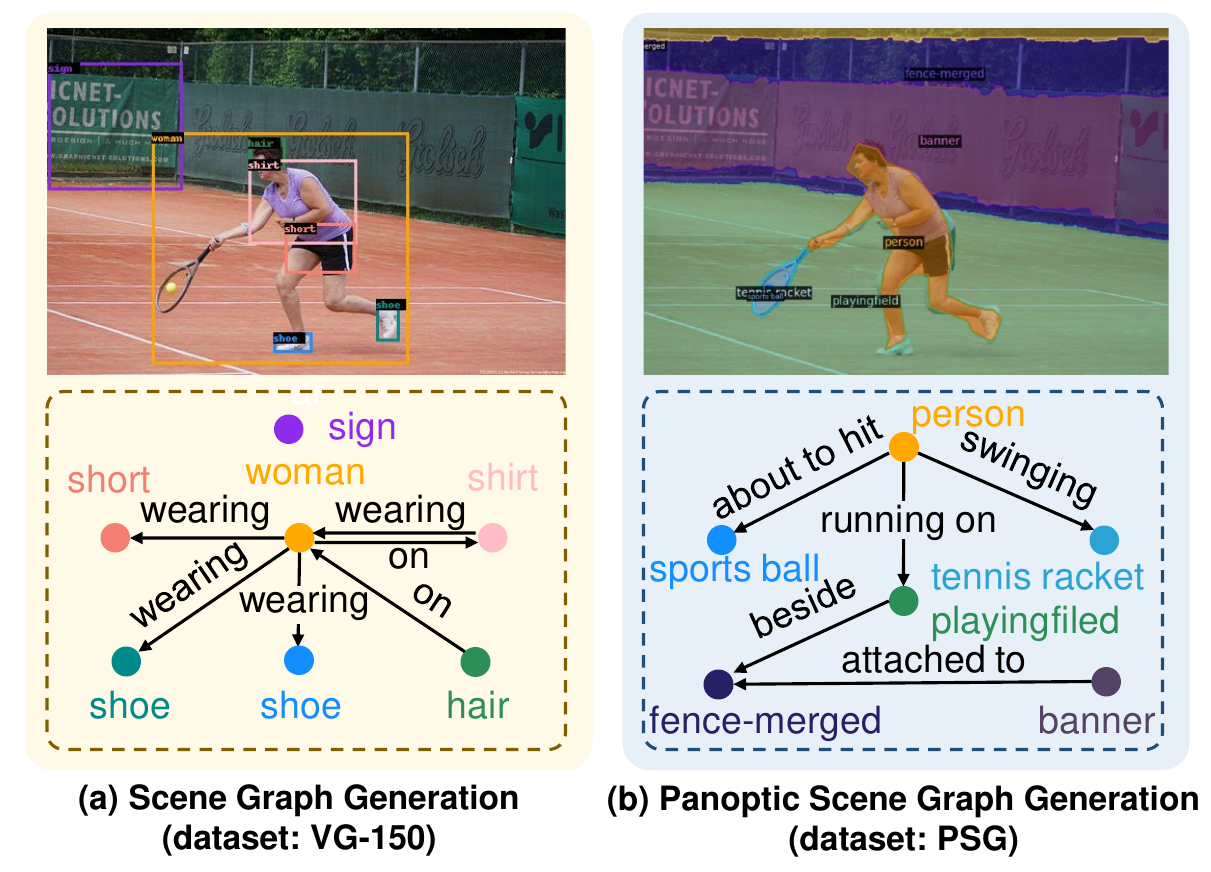}
    \vspace{-1.0em}
  \caption{(a) Scene Graph Generation (SGG): It relies on the bounding box-based paradigm, which can lead to inaccurate object localization and limited background annotation. (b) Panoptic Scene Graph Generation (PSG): It presents a more comprehensive and cleaner scene representation, with more accurate localization of objects and including relationships with the background (\eg, \texttt{fence} and \texttt{playingfield}).}
  \label{fig:difference}
\end{figure}

Scene Graph Generation (SGG)~\cite{xu2017scene,chen2019counterfactual} is a fundamental scene understanding task~\cite{zhou2019semantic,li2023learning,DBLP:journals/pami/TangSQLWYJ17,chen2024neural} that surpasses mere object classification and localization by predicting relations between objects in a scene~\cite{zhao2023unified,zareian2020learning,li2023compositional,yu2021cogtree}. However, due to its reliance on the bounding box-based paradigm, traditional SGG suffers from inaccurate object localization and limited background annotation. To address these issues, a novel variant of SGG called Panoptic Scene Graph Generation (\textbf{PSG})~\cite{yang2022panoptic} has emerged. As shown in Fig.~\ref{fig:difference}, PSG leverages more fine-grained scene mask representations (\ie, panoptic segmentation) and defines relations for background stuff (\eg, \texttt{playingfiled}), thus providing a more precise and comprehensive understanding of the scene~\cite{DBLP:conf/nips/Li0CSZC23,DBLP:journals/corr/abs-2401-03890,DBLP:journals/pami/LiTM19}.

\begin{figure*}[t]
  \centering
  \includegraphics[width=\linewidth]{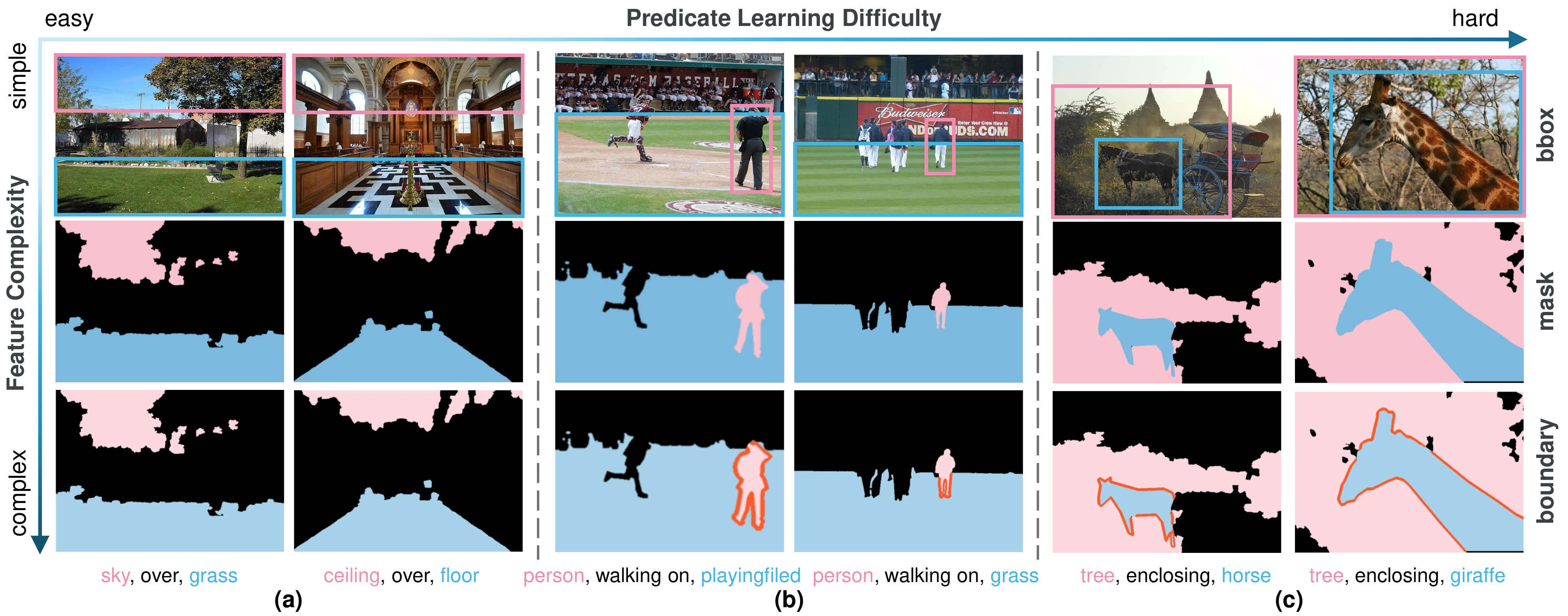}
  \vspace{-1.0em}
  \caption{
With the escalation of learning difficulty in predicates, there is a corresponding increase in the complexity of features necessary for accurately predicting pairwise relations between objects. These features include traditional bbox features and our proposed shape-aware features (\ie, mask and boundary features).}
  \label{fig:intro}
\end{figure*}

Although PSG has made notable progress \cite{yang2022panoptic,li2023panoptic,zhou2023hilo}, almost all existing approaches draw inspiration from the strategies established in SGG~\cite{zheng2023prototype,chen2023addressing,li2022devil,li2022integrating}. Unfortunately, almost all existing approaches overlook the importance of \textbf{\emph{Shape-aware Features}}, which inherently concentrate more on the contours and boundaries of objects\footnote{In this paper, we refer to both \emph{things} and \emph{stuff} as objects. \label{footnote:object}}. To be more specific, state-of-the-art PSG methods just replace object features with better representations from panoptic segmentors, and they all still utilize spatial features derived from the minimum bounding boxes (bbox) of these masks, neglecting the critical shape-aware features. This limitation can hinder a holistic grasp of the scene, potentially resulting in semantic confusion in fine-grained visual relation prediction. Take \texttt{person-playingfiled} in Fig.~\ref{fig:intro}(b) as an example, the similarity in their bbox-based spatial features can easily lead to confusion between relations like \texttt{walking on} and \texttt{running on}. 

To this end, we argue that it is essential to incorporate shape-aware features into PSG rather than relying solely on spatial features based on bboxes. By ``shape-aware features'', we mainly mean two types of features: 1) \textbf{Mask Features}: These features exploit the details encompassed in fine-grained mask representations, including the shape and contour of the objects. This inclusion captures a wealth of visual intricacies that significantly enhance the accuracy of relation prediction. As shown in Fig.~\ref{fig:intro}(b), for predicates representing actions like \texttt{walking on}, there is a commonality in the shape and contour of the subject (\eg, \texttt{person}). In such scenarios, by using mask features, even when faced with identical subject-object pairs (\eg, \texttt{person-playingfiled}), the model can still alleviate semantic confusion and disambiguate relationships (\eg, \texttt{running on} and \texttt{standing on}). 2) \textbf{Boundary Features}: These features are extracted from the intersection of subject and object masks, which provide unique advantages in cases of interaction and contact between subject-object pairs. When dealing with predicates like \texttt{enclosing} (\cf, Fig.~\ref{fig:intro}(c)), the mask features of separate subjects (\eg, \texttt{tree}) or objects (\eg, \texttt{horse}) display a range of diversity. This diversity can result in semantic confusion, rendering the use of mask features alone inadequate for making accurate predictions. However, there is a notable resemblance (\eg, encirclement) between \texttt{tree-horse} and \texttt{tree-giraffe}. Thus, the incorporation of boundary features that represent the intersection of subject-object pairs can aid in enhancing prediction accuracy.

Drawing inspiration from cognitive psychology research~\cite{sarafyazd2019hierarchical} that indicates humans tend to learn concepts progressively, starting from easier concepts and gradually advancing to comprehend harder ones, we propose to integrate shape-aware features in an easy-to-hard manner. Specifically, as the difficulty of learning predicates increases, we progressively enhance the complexity of features. For example, certain simple positional relations like \texttt{over} (\cf, Fig.~\ref{fig:intro}(a)) can be accurately predicted using traditional bbox features alone. For more complex relations like \texttt{walking on} (\cf, Fig.~\ref{fig:intro}(b)), relying solely on bbox features can lead to semantic confusion, while incorporating mask features can help disambiguate the relationships. When facing more challenging predicates like \texttt{enclosing} (\cf, Fig.~\ref{fig:intro}(c)), accurate predictions heavily rely on the effective utilization of boundary features, which capture the interaction between subject-object pairs.

In this paper, we propose a novel Curricular shApe-aware FEature (\textbf{CAFE}) learning strategy for PSG. CAFE is a model-agnostic strategy which skillfully weaves into the training process via a curriculum learning strategy. Specifically, we first categorize the predicates into three groups based on cognitive difficulty, \ie, predicate distribution and semantic diversity. Then, we divide the training process into three stages, with each stage utilizing its own relation classifier. These classifiers are tailored to handle predicates with increasing learning difficulties and are equipped with corresponding sets of features of ascending complexity (\ie, bbox features, mask features, and boundary features). We also incorporate knowledge distillation~\cite{li2023label,li2022nicest} to retain knowledge acquired in earlier stages.

We conducted comprehensive experiments on the challenging PSG dataset~\cite{yang2022panoptic}, exploring both robust PSG and zero-shot PSG. Since CAFE is model-agnostic, it can be seamlessly incorporated into any advanced PSG architecture\footnote{Considering that our shape-aware features originate from masks generated by the panoptic segmentor and taking into account our limited computational resource, it is not straightforward to apply CAFE to one-stage methods. The ``PSG architecture" here represents all two-stage frameworks.} and consistently improves its performance. In robust PSG, our method can obtain a new state-of-the-art performance on the trade-off between different metrics.  Extensive ablations and results on multiple PSG tasks have shown the robustness and generalization capabilities of the CAFE. Moreover, in zero-shot PSG, our CAFE can infer unseen visual relation triplets by capitalizing on the robust visual relation features learned during training.

In summary, we make three contributions in this paper:

\begin{enumerate}[leftmargin=4mm]
    \item We delve deeply into the PSG task and unveil a critical issue: the exclusive reliance on spatial features based on bboxes while disregarding shape-aware features. 
    
    \item We propose the model-agnostic curricular feature training with shape-aware feature preparation for PSG, which allows the model to learn in an easy-to-hard manner.
    
    \item Extensive results show the robustness and effectiveness of CAFE, \ie, it outperforms existing state-of-the-art methods by a large margin on the PSG benchmark.
\end{enumerate}

\section{Related Work} \label{sec2}

\noindent\textbf{Panoptic Scene Graph Generation (PSG).} PSG aims to transform an image into a structured graph representation, formulated as a series of visual relation triplets. Contrary to SGG~\cite{DBLP:conf/cvpr/JungKKC23,DBLP:journals/pami/LyuGZSS23,DBLP:conf/iccv/Yu00T0Z23,DBLP:conf/iccv/LuRCKY0TV21}, PSG not only employs a more fine-grained scene representation but also addresses the challenge of missing background context. Existing PSG models can be divided into two groups: 1) Two-stage PSG: They first utilize a pretrained panoptic segmentation model (\eg, Panoptic FPN~\cite{kirillov2019panoptic} or Panoptic Segformer~\cite{li2022panoptic}) to generate masks and then predict the classes of objects and their pairwise relations~\cite{li2023panoptic,jin2023fast}. This paradigm allows classic SGG models~\cite{xu2017scene, lin2020gps,li2021bipartite} to be adapted with minimal modifications. 2) One-stage PSG: These models construct an end-to-end model to detect the objects and relations from image features directly~\cite{yang2022panoptic,zhou2023hilo,DBLP:journals/corr/abs-2307-08699}. In this paper, we build upon two-stage baselines and propose a model-agnostic approach that can be incorporated into any PSG model.

\noindent\textbf{Shape-Aware Features for Vision Tasks.} Shape-aware features are a type of visual information representation that place a stronger emphasis on the shape of objects within an image, which benefits various vision tasks~\cite{tian2019learning,cao2021shapeconv,gomes2021shape,jinka2023sharp}. Among these features, boundary-aware features play a pivotal role in enhancing the understanding of object boundaries. For example, in semantic segmentation tasks~\cite{DBLP:conf/nips/ZhangZT0S20}, several methods have been proposed to incorporate boundary-aware information, including feature propagation~\cite{ding2019boundary}, geometric encoding~\cite{gong2021boundary}, and graph convolution~\cite{hu2021boundary}. Unlike these methods, we are the first approach that leverages shape-aware features to represent interaction information between objects, enhancing relation prediction.

\noindent\textbf{Curriculum Learning (CL).} Curriculum learning~\cite{bengio2009curriculum, jiang2015self,wang2021survey} is a training strategy that trains the model from easier data to harder data, which mimics the human recognition process~\cite{soviany2022curriculum,guo2018curriculumnet,yuan2022easy}. 
It has been demonstrated to significantly enhance performance across a variety of machine learning tasks~\cite {zhang2019leveraging,yang2023dynamic,wang2023hierarchical,xu2023dynamic}. Moreover, when it comes to imbalanced data, CL is also an effective debiasing strategy. For instance, DCL~\cite{wang2019dynamic} achieves strong discrimination through adaptive adjustments of the sampling strategy and loss weights. Besides, CLDL~\cite{li2023curriculum} effectively discerns easy and hard samples by decomposing the segmentation task into multiple label distribution estimation tasks. Compared to these applications, PSG is a sophisticated task that involves predicates of varying learning complexities. Our CAFE gauges cognitive difficulty using long-tailed predicate distribution and semantic similarity. Inspired by human cognition, we employ CL to progressively integrate shape-aware features in an easy-to-hard manner, enhancing feature complexity with increasing difficulty. 

\begin{figure*}[!htbp]
  \centering
  \includegraphics[width=\linewidth]{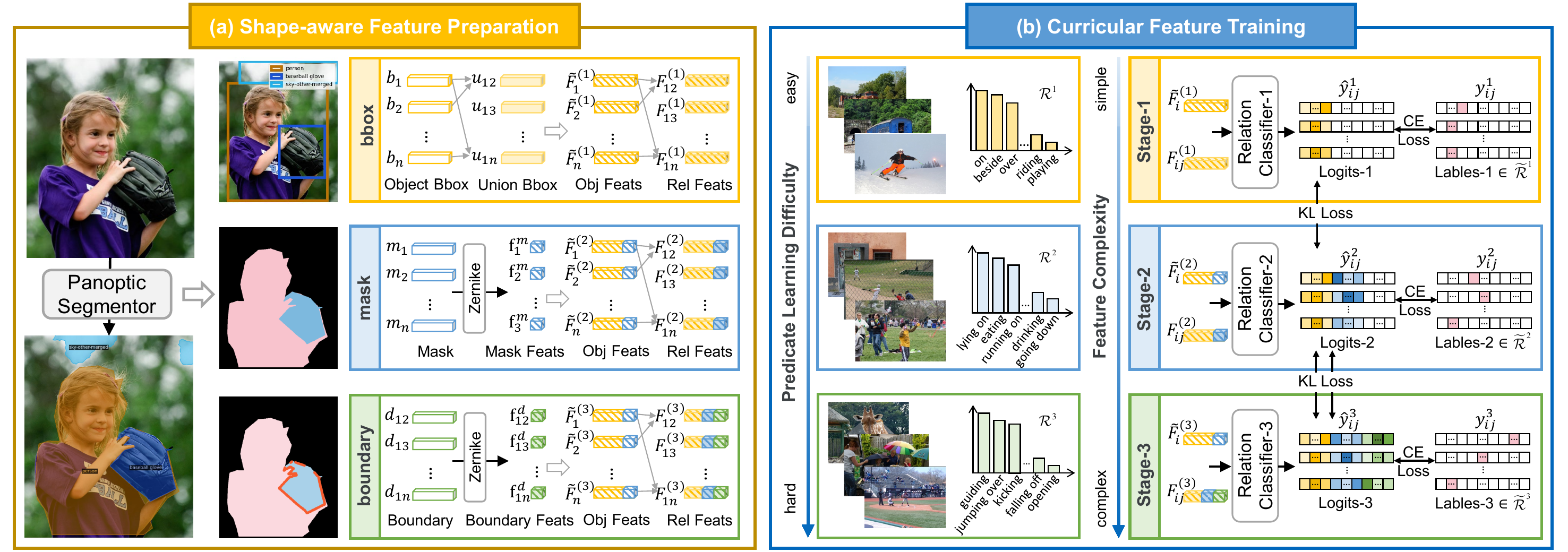}
  \vspace{-1.0em}
  \caption{The pipeline of our proposed CAFE. (a) Shape-aware Feature Preparation: we generate three types of features, ranging in complexity from simple to complex. (b) Curricular Feature Learning: we divide the training process into three stages according to the predicate learning difficulty from easy to hard. Each stage utilizes the corresponding features and has its own relation classifier. The training objectives include cross-entropy (CE) loss and KL loss.}
  \label{fig:pipeline}
\end{figure*}

\section{Approach}\label{sec3}
\noindent\textbf{Problem Formulation.}
PSG task aims to generate a panoptic scene graph $\mathcal{G}$ for a given image $\mathcal{I} \in \mathbb{R}^{H \times W \times 3}$. The scene graph consists of a set of nodes $\mathcal{N}$ and a set of edges $\mathcal{E}$, denoted as $\mathcal{G} = \{\mathcal{N} = \{o_i; m_i\}; \mathcal{E} = \{r_{ij}\}\}$. Each object\footref{footnote:object} is represented by a binary mask $m_i \in \mathcal{M}$ associated with an object category $o_i \in \mathcal{O}$. The relation category between the $i$-th and $j$-th objects is denoted by $r_{ij} \in \mathcal{R}$. $\mathcal{M}$, $\mathcal{O}$, and $\mathcal{R}$ represent the sets of all object masks, object categories, and relation categories, respectively. Besides, the binary masks $m_i \in \{0,1\}^{H \times W}$ do not overlap, \ie, $\sum_{i=1}^{n} m_i \leq \mathbf{1}^{H \times W}$. Hence, the PSG task models the following distribution:
\begin{equation}
	\label{E:psg_def}
	P(\mathcal{G}| \mathcal{I})=P(\mathcal{M},\mathcal{O},\mathcal{R}|\mathcal{I}).
\end{equation}
\subsection{Overview: Two-Stage PSG Approach}\label{3.1}
Similar to SGG, PSG has two-stage and one-stage baselines. In this paper, we build upon the two-stage framework and propose a model-agnostic approach to enhance the performance. A typical two-stage PSG model involves three steps: mask generation, object classification, and relation classification. Thus, the PSG task $\operatorname{P}(\mathcal{G}|\mathcal{I})$ is decomposed into:
\begin{equation}
	\label{E:psg_twostage_def}
	P(\mathcal{G}| \mathcal{I})=P(\mathcal{M}| \mathcal{I}) \cdot {P}(\mathcal{O} | \mathcal{M}, \mathcal{I}) \cdot P(\mathcal{R} | \mathcal{O}, \mathcal{M}, \mathcal{I}).
\end{equation}
\noindent\textbf{Mask Generation $P(\mathcal{M}| \mathcal{I})$.} This step aims to segment an image into a set of masks $\mathcal{M}$ with panoptic segmentation.

\noindent\textbf{Object Classification ${P}(\mathcal{O} | \mathcal{M}, \mathcal{I})$.} This step predicts the object category of each $m_i \in \mathcal{M}$. It consists of an object context encoder $\mathtt{Enc}_{obj}$ to extract the object feature $F_i$ and an object classifier $\mathtt{Cls}_{obj}$ to predict the object categories $\widehat{o}_i$.

\noindent\textbf{Relation Classification $P(\mathcal{R} | \mathcal{O}, \mathcal{M}, \mathcal{I})$.} This step predicts the relation of every two masks in $\mathcal{M}$ along with their object categories in $\mathcal{O}$. It comprises a relation context encoder $\mathtt{Enc}_{rel}$ and a relation classifier $\mathtt{Cls}_{rel}$. The former performs context modeling to extract refined object features $\widetilde{F}_i$ for object classification using $\mathtt{Cls}_{obj}$, and also conducts relation feature encoding (\cf, Eq.~\ref{eq6}) for relation prediction with $\mathtt{Cls}_{rel}$. The latter takes relation features $F_{ij}$ as input to predict the relation distribution.

Our proposed CAFE consists of shape-aware feature preparation (\cf, Sec.~\ref{3.2}) and curricular feature learning (\cf, Sec.~\ref{3.3}). The pipeline is illustrated in Fig.~\ref{fig:pipeline}.

\subsection{Shape-aware Feature Preparation}
\label{3.2}

To address the semantic confusion issue arising from relying solely on bbox features, we propose two types of shape-aware features, \ie, mask features and boundary features. Afterward, we propose the stage-wise feature fusion strategy to obtain three stages of features (\cf, Fig.~\ref{fig:pipeline}(a)), which are used for the relation classification step.

\subsubsection{Shape-aware Features Extraction} 
\label{3.2.1}
Shape-aware features extraction is used to extract two types of shape-aware features: 1) \emph{mask features}: they focus on the shape and contour of objects, providing richer visual information. 2) \emph{boundary features}: they are derived from the intersection of subject and object masks, capturing the interactions between object pairs in the scene. 

For the mask features, we employ binary erosion operation to extract the object contour from the binary mask representation $m_i \in \{0,1\}^{H \times W}$. Then, to obtain a compact representation of the shape information, we adopt Zernike moments~\cite{von1934beugungstheorie,noll1976zernike} to transform the extracted contour into mask feature  $f^{m}_{i} \in \mathbb{R}^{256}$. The calculation process is as follows:
\begin{equation}
    f^{m}_{i} = \mathtt{Zk}(m_i \cap (\neg\mathtt{ero}(m_i))),
\end{equation}
where $\mathtt{Zk}(\cdot)$ denotes the computation of Zernike moments, and $\mathtt{ero}(\cdot)$ denotes the binary erosion operation. 

For the boundary features, we first calculate the intersection $d_{ij}$ between the subject mask $m_i$ and the object mask $m_j$. Then, similar to the mask features, we calculate boundary features $f^{d}_{ij} \in \mathbb{R}^{256}$ as follows:
\begin{equation}
    d_{ij} = m_i \cap m_j, f^{d}_{ij} = \mathtt{Zk}(d_{ij} \cap (\neg \mathtt{ero}(d_{ij}))).
\end{equation}
Furthermore, in the absence of an interaction area between the subject and object, signifying no overlap between the subject mask and the object mask, the vector $f^{d}_{ij}$ transforms into a zero vector.

\subsubsection{Stage-wise Feature Fusion} 
\label{3.2.2}
This step generates three stages of features with different complexities, based on the difficulty of learning predicates. These stages of features are used in the subsequent object and relation classification step, and they can be subdivided into refined object features $\widetilde{F}_i$ and relation features $F_{ij}$.

\emph{First Stage.} Since the predicates are relatively straightforward to learn, only simple bbox features are needed. To obtain these features, we calculate the tightest bounding box $b_i$ of the mask $m_i$ obtained in the mask generation step. Subsequently, we use the RoIAlign operation~\cite{he2017mask} to extract the visual feature $v_i$. With $b_i$ and $v_i$ as input, we derive the object feature $F^{\left(1\right)}_i$ through an object context encoder $\mathtt{Enc}^{\left(1\right)}_{obj}$. Then, we perform context modeling to extract the refined object feature $\widetilde{F}^{\left(1\right)}_i$ as follows:
\begin{equation}
    F^{\left(1\right)}_i = \mathtt{Enc}^{\left(1\right)}_{obj}(v_i \oplus b_i), \widetilde{F}^{\left(1\right)}_i = \mathtt{Enc}^{\left(1\right)}_{rel}(v_i \oplus F^{\left(1\right)}_i \oplus w_i),
\end{equation}
where $\oplus$ denotes concatenation, and $w_i$ is the GloVe embedding of object class $\widehat{o}_i$. Afterwards, we combine the pairwise refined object features with the union feature to obtain the relation features $F^{(1)}_{ij}$ between the $i$-th and $j$-th objects:
\begin{equation}
    F^{\left(1\right)}_{ij} = (\widetilde{F}^{\left(1\right)}_i \oplus \widetilde{F}^{\left(1\right)}_j) \otimes u_{ij},
\label{eq6}
\end{equation}
where $\otimes$ denotes element-wise multiplication, and $u_{ij}$ denotes the visual feature of the union box $b_{ij}$.

\emph{Second Stage.} As the predicates become more complex, we incorporate mask features to disambiguate the relationships. To achieve this, we combine the original bbox feature $f^b_i$ with the mask feature $f^m_i$, and then utilize a relation encoder $\mathtt{Enc}^{\left(2\right)}_{rel}$ to obtain the object feature $\widetilde{F}^{\left(2\right)}_i$:
\begin{equation}
    \widetilde{F}^{\left(2\right)}_i = \mathtt{Enc}^{\left(2\right)}_{rel}(v_i \oplus F^{\left(1\right)}_i \oplus w_i \oplus f^m_i).
\end{equation}
The way of obtaining relation features $F^{\left(2\right)}_{ij}$ is similar to the first stage.

\emph{Third Stage.} In order to capture interactions between subject-object pairs, we incorporate boundary features to enhance the relation features. The method of obtaining object features remains the same as in the second stage. However, for relation features $F^{\left(3\right)}_{ij}$, we combine the union feature with the boundary feature $f^d_{ij}$ to better capture the interaction between object pairs. The calculation is as follows:
\begin{equation}
    F^{\left(3\right)}_{ij} = (\widetilde{F}^{\left(2\right)}_i \oplus \widetilde{F}^{\left(2\right)}_j) \otimes (u_{ij} \oplus f^d_{ij}).
\end{equation}
\begin{figure}[!t]
  \centering
  \includegraphics[width=\linewidth]{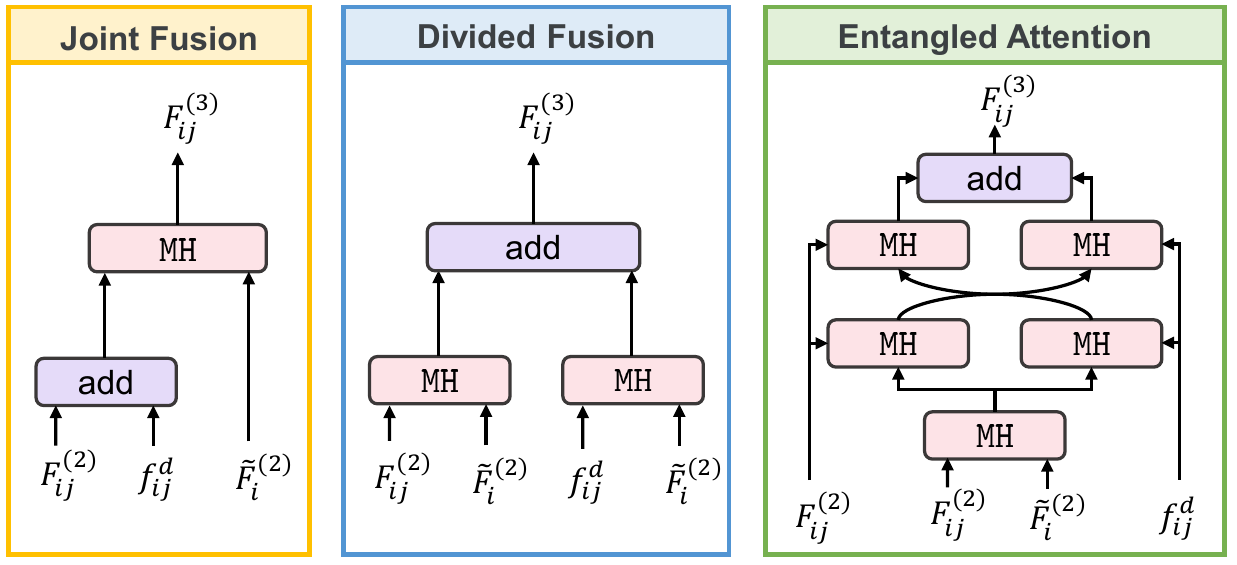}
   \vspace{-1.0em}
  \caption{Different instantiations of feature fusion strategies in the Transformer-based CAFE.} 
  \label{fig:fusion}
\end{figure}
\indent Furthermore, for the Transformer-based CAFE, we draw insights from~\cite{DBLP:conf/emnlp/CaoCSZZ21,li2019entangled} to fully leverage the potential of the Transformer in enhancing model performance. 
We delve deeply into the attention mechanism and devise three additional feature fusion strategies. These feature fusion strategies are built upon concatenation, differing solely in the integration of boundary features (\ie, the generation of $F^{\left(3\right)}_{ij}$). Fig.~\ref{fig:fusion} illustrates the process of generating $F^{\left(3\right)}_{ij}$ with different feature fusion strategies.

\emph{Joint Fusion.} Before performing attention computation, it incorporates the boundary feature $f^{d}_{ij}$ along with the second-stage relation feature $F^{\left(2\right)}_{ij}$. Then, we use the multi-head attention function $\mathtt{MH} (\cdot, \cdot,\cdot)$~\cite{vaswani2017attention} to generate the third-stage relation feature: 
\begin{equation}
F^{\left(3\right)}_{ij}=\mathtt{MH}(F^{\left(2\right)}_{ij}+f^{d}_{ij},\widetilde{F}^{\left(2\right)}_i,\widetilde{F}^{\left(2\right)}_i).
\end{equation}

\emph{Divided Fusion.} It decomposes the shape-aware feature fusion into two parallel branches, with the final results being summed up. The calculation is as follows: 
\begin{equation}
F^{\left(3\right)}_{ij}=\mathtt{MH}(F^{\left(2\right)}_{ij},\widetilde{F}^{\left(2\right)}_i,\widetilde{F}^{\left(2\right)}_i)+\mathtt{MH}(f^{d}_{ij},\widetilde{F}^{\left(2\right)}_i,\widetilde{F}^{\left(2\right)}_i).
\end{equation}
\indent\emph{Entangled Attention.} It fuses boundary features in an entangled manner, which can enhance the ability to leverage the complementary nature of different features during attention operations. Specifically, we first feed object feature $F^{\left(2\right)}_i$ and relation feature $F^{\left(2\right)}_{ij}$ into a multi-head self-attention layer to generate the attention map $A$. This output $A$ is then integrated into the entangled attention along with mask and boundary features. The calculation is as follows:
\begin{equation}
    A = \mathtt{MH}(F^{\left(2\right)}_{ij}, \widetilde{F}^{\left(2\right)}_i, \widetilde{F}^{\left(2\right)}_i).
\end{equation}
\indent Then, we use the multi-head attention function to generate preliminary information $I^m$ and $I^d$ of the mask feature and boundary feature:
\begin{equation}
    I^m = \mathtt{MH}(A, F^{\left(2\right)}_{ij}, F^{\left(2\right)}_{ij}), I^d = \mathtt{MH}(A, f^d_{ij}, f^d_{ij}).
\end{equation}
Finally, we utilize $I^d$ and $I^m$ as an effective guidance in the calculation of mask and boundary attention. The relation feature of the third stage $F^{\left(3\right)}_{ij}$ is calculated as follows:
\begin{equation}
    F^{\left(3\right)}_{ij} = \mathtt{MH}(I^d, F^{\left(2\right)}_{ij}, F^{\left(2\right)}_{ij}) + \mathtt{MH}(I^m, f^d_{ij}, f^d_{ij}).
\end{equation}
\subsection{Curricular Feature Training}
\label{3.3}

Given prepared shape-aware features, our goal is to incorporate these features into training, enabling the model to learn in an easy-to-hard manner(\cf, Fig.~\ref{fig:pipeline}(b)). To achieve this, we introduce a cognition-based predicate grouping strategy to categorize the predicates into three reasonable groups. Subsequently, we configure the classification space for the relation classifiers and design a sampling strategy to achieve a relatively balanced group.

\subsubsection{Cognition-based Predicate Grouping} 
\label{3.3.1}
Cognition-based predicate grouping categorizes the predicate set into three mutually exclusive groups based on the varying levels of learning difficulty. Recognizing that biased data distribution and semantic similarity can potentially hinder recognition capability, we aim to ensure each group maintains a relative balance and avoids the inclusion of semantically similar predicates. 

\begin{figure}[!t]
    \centering
    \includegraphics[width=1.0\linewidth]{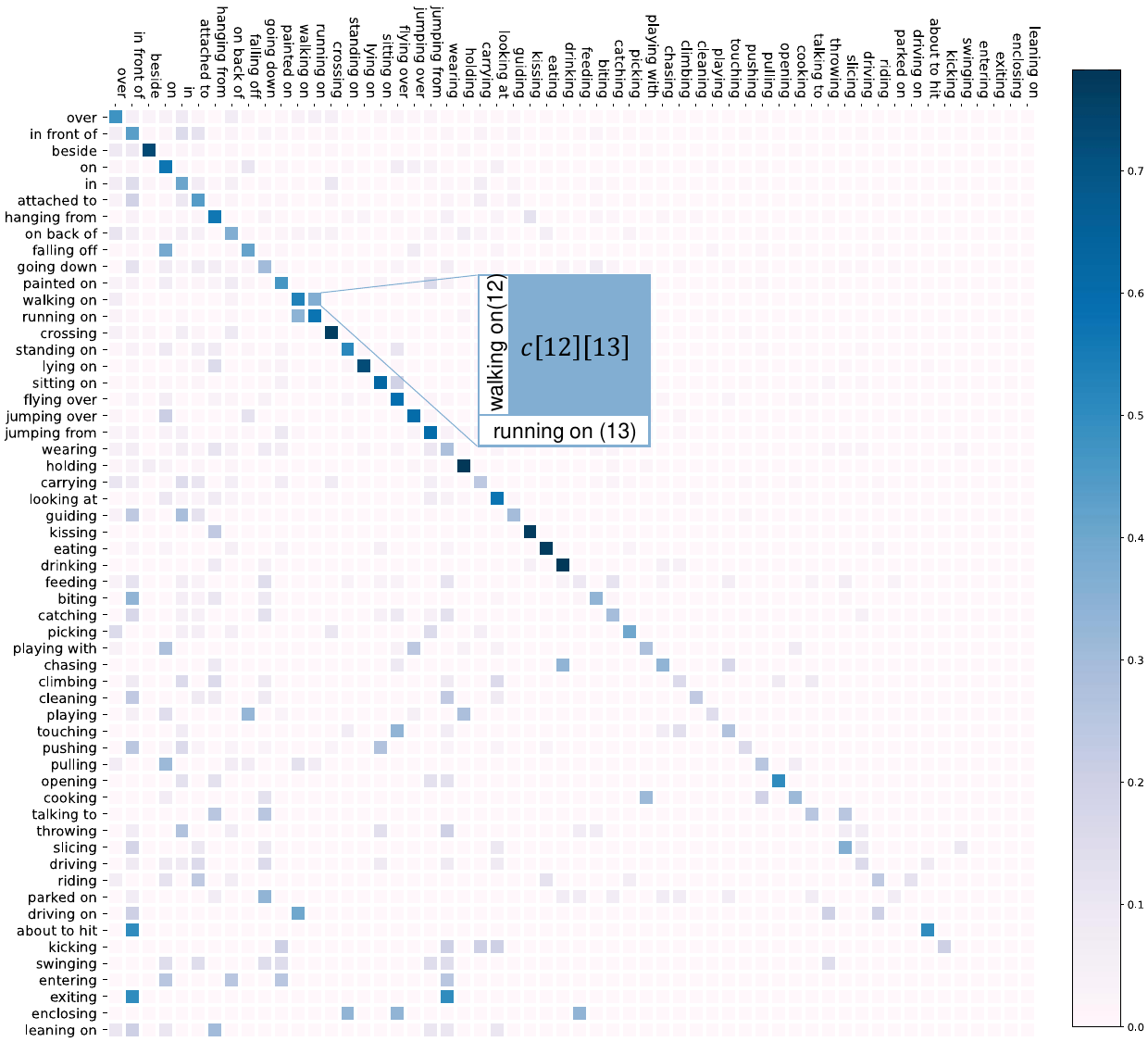}
    \caption{A confusion matrix of our Motifs+CAFE model. The element $\mathcal{C}[r_i][r_j]$ means the number of samples labeled as predicate $r_i$ but predicted as $r_j$. For instance, $\mathcal{C}[12][13]$ corresponds to the number of instances where the GT label is ``\texttt{walking on}", but the prediction was``\texttt{running on}".}
    \label{fig:confusion matrix}
\end{figure}

Our cognition-based predicate grouping strategy comprises two steps: 1) \emph{Predicate Distribution-based Grouping}: We first obtain a sorted predicate set $\mathcal{R}$ by sorting the predicate classes in descending order based on the amount of training instances. Then, we evenly distribute the predicates into three distinct groups, denoted as ${\mathcal{R}^{1}, \mathcal{R}^{2}, \mathcal{R}^{3}}$. By uniformly dividing the sorted predicate set into three groups, we can reduce the ratio between the most and least annotated predicates from 6,777 to 28, even for the most imbalanced group. Besides, it's worth noting that the learning difficulty of predicates in these three groups are arranged from easy to hard. 2) \emph{Semantic Similarity-based Adjustment}: We first compute the confusion matrix $\mathcal{C}$ (\cf, Fig.~\ref{fig:confusion matrix}), where each element $\mathcal{C}[i][j]$ means the number of samples labeled as the $i$-th relation category but predicted as the $j$-th one. We also perform normalization on the values in the confusion matrix, ensuring that the sum of each row is equal to 1. Then, we assess each group for substantial semantic similarity (\ie, confusion matrix value $\mathcal{C}[i][j]$ exceeds the similarity threshold $\mu$) and make corresponding adjustments. If detected within the $k$-th group $\mathcal{R}^{k}$, we either move the confusing predicate with lower occurrence to $\mathcal{R}^{k+1}$ or relocate predicate with higher occurrence to $\mathcal{R}^{k-1}$. For instance, following the initial grouping, \texttt{riding} and \texttt{driving}) remain in the same group $\mathcal{R}^{1}$. Then, we relocate \texttt{driving}, which has a lower frequency, to $\mathcal{R}^{2}$. The specific implementation is displayed in Algorithm~\ref{alg:algorithm1}.

\begin{algorithm}[t]
	\caption{Cognition-based Predicate Grouping}
	\label{alg:algorithm1}
	\KwIn{Sorted predicate set $\mathcal{R}=\{r_{i}\}_{i=1}^N$, prediction confusion matrix $\mathcal{C} \in \mathbb{R}^{N \times N}$, predicate number $N$, similarity threshold $\mu$, and predicate group number $K$.}
	\KwOut{Three mutually exclusive groups $\{\mathcal{R}^{1}, \mathcal{R}^{2}, \mathcal{R}^{3}\}$.}
	\BlankLine
    \tcc{Initialize each predicate group as an empty set}
    \For{$i \gets 1$ \KwTo $K$}{
      ${R}^{i} = \{\}$
    }
    
    \tcc{Step 1: Predicate Distribution-based Grouping}
    
    $n \gets N \div K$\;
    \For{$k \gets 1$ \KwTo $K-1$}{
        \For{$i \gets (k-1) \cdot n$ \KwTo $k \cdot n - 1$}{
            Add $r_i$ to $\mathcal{R}^{k}$\;
        }
    }

    \For{$i \gets (K-1) \cdot n$ \KwTo $N-1$}{
        Add $r_i$ to $\mathcal{R}^{K}$\;
    }
    \tcc{Step 2: Semantic Similarity-based Adjustment}
    
    \If{$k = 1$ or $k = 2$}{
        \For{$i \gets 1$ \KwTo $\text{len}(\mathcal{R}^{k}) - 1$}{
            \For{$j \gets i + 1$ \KwTo $\text{len}(\mathcal{R}^{k})$}{
                \If{$\mathcal{C}[r_i][r_j] \geq \mu$}{
                Move $r_j$ from $\mathcal{R}^{k}$ to $\mathcal{R}^{k+1}$\;
                }
            }
        }
    }
    \Else{
        \For{$i \gets 1$ \KwTo $\text{len}(\mathcal{R}^{k}) - 1$}{
            \For{$j \gets i + 1$ \KwTo $\text{len}(\mathcal{R}^{k})$}{
            \If{$\mathcal{C}[r_i][r_j] \geq \mu$}{
                Move $r_i$ from $\mathcal{R}^{k}$ to $\mathcal{R}^{k-1}$\;
            }
            }
        }
    }
 
\label{alg:1}
\end{algorithm}

\subsubsection{Classification Space Configuration}
\label{3.3.2}
Classification space configuration aims to configure the classification space for the relation classifiers. To be specific, we employ the curriculum learning strategy and utilize three relation classifiers denoted as $\mathtt{Cls}^{\left(k\right)}_{rel}$, where $k \in {1, 2, 3}$. These classifiers are designed to follow a continuously growing classification space. Each classifier, except for the $\mathtt{Cls}^{\left(1\right)}_{rel}$, is designed to recognize predicate classes from both previous and current groups. The classification space in $\mathtt{Cls}^{\left(k\right)}_{rel}$ is defined as $\widetilde{\mathcal{R}}^k = \mathcal{R}^1 \cup \cdots \cup \mathcal{R}^k$.

\begin{figure*}[!htbp]
    \centering
    \includegraphics[width=1.0\linewidth]{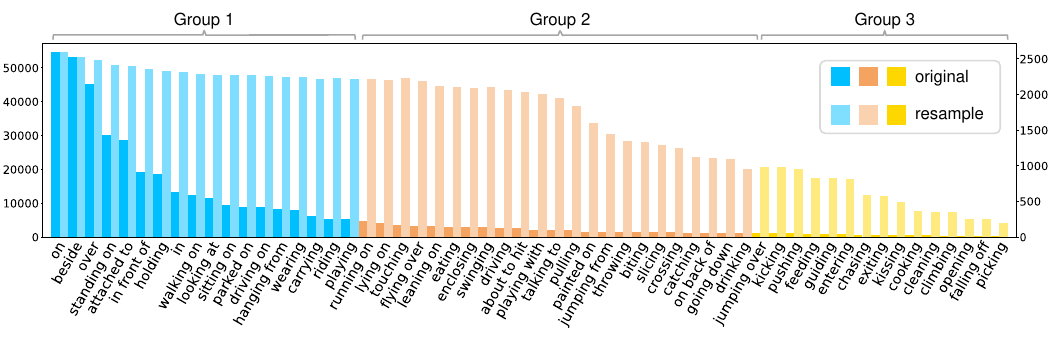}
    \vspace{-1.0em}
    \caption{The final results of our cognition-based predicate grouping and predicate sampling strategy. We categorize the predicates into three mutually exclusive groups $\mathcal{R}^1$ (\textcolor[RGB]{0,191,255}{blue}), $\mathcal{R}^2$ (\textcolor[RGB]{244,164,96}{orange}), $\mathcal{R}^3$ (\textcolor[RGB]{255,215,0}{yellow}), based on the cognition difficulty. For each predicate, the left bar represents the numbers from the original data annotations, while the right bar represents the numbers after predicate sampling.}
    \label{fig:grouping}
\end{figure*}

\subsubsection{Predicate Sampling} 
\label{3.3.3}

Due to severe dataset imbalance, the model exhibits a bias towards head predicates. Correspondingly, it is necessary to provide each relation classifier with a more balanced training set. To achieve a decent balance for subsequent optimization, we employ a two-fold approach. 

Firstly, for the entire dataset, we conduct oversampling on images containing tail predicate samples, thereby augmenting the abundance of tail predicate samples. Secondly, to mitigate the significant variance in the number of triplets among different predicates within the same group, we draw inspiration from the median resampling strategy~\cite{dong2022stacked} and design a sampling rate $\phi_k$. Specifically, within each classification space of classifier, we compute the median value denoted as $Med$, as well as the maximum value denoted as $Max$, across all predicates within $\widetilde{\mathcal{R}}^k$. Subsequently, for each predicate $r_k$ within $\widetilde{\mathcal{R}}^k$, we formulate two sampling rates: $\phi_1^k$ and $\phi_2^k$. The former is biased towards the head predicates, while the latter is skewed towards the tail predicates. The calculation of $\phi_1^k$ and $\phi_2^k$ is as follows:
\begin{equation}
\phi_1^k =\left\{
\begin{aligned}
&  \frac{Cnt(r_k)}{Med}&, if\ Cnt(r_k) \ge Med, \\
&  \qquad 1  &, if \ Cnt(r_k) < Med. \\
\end{aligned}
\right.
\end{equation}
\begin{equation}
\phi_2^k =\left\{
\begin{aligned}
&  \frac{Max\! -\! Cnt(r_k)}{Max - Med}\!&\!, if\  Cnt(r_k) \ge Med, \\
&  \frac{Max\! -\! Cnt(r_k)}{Max - Med} * k ^2\!&\!, if\  Cnt(r_k) < Med. \\
\end{aligned}
\right.
\end{equation}
\indent In order to strike a trade-off between the two sampling rates, we introduce a hyperparameter $\lambda$ to balance $\phi_1^k$ and $\phi_2^k$. The final sampling rate $\phi_k$ is calculated as follows:
\begin{equation}
   \phi_k = \lambda  \phi_1^k + (1-\lambda) \phi_2^k.
\end{equation}
\indent By employing these two sampling operations, the training sets within each classifier achieve a relatively balanced distribution. This facilitates the learning of discriminative representations towards the designated subsets of predicates. The results of cognition-based predicate grouping and predicate sampling are illustrated in Fig.~\ref{fig:grouping}.

\subsection{Training Objectives and Inference}

\noindent\textbf{Relation Prediction.} In this step, the fused features are transformed into relation distributions (\cf, Fig.~\ref{fig:pipeline}(b)), and the classifiers are optimized using cross-entropy loss. Firstly, we input the fused relation features $F^{\left(k\right)}_{ij}$ from the $k$-th stage into the corresponding relation classifier $\mathtt{Cls}^{\left(k\right)}_{rel}$ to obtain the $k$-th relation distribution $\widehat{y}^k_{ij}$:
\begin{equation}
    \widehat{y}^k_{ij} = \mathtt{Cls}^{\left(k\right)}_{rel}(F^{\left(k\right)}_{ij}).
\end{equation}
\indent Subsequently, we optimize these three relation classifiers simultaneously, and the training objective function is calculated as follows:
\begin{equation}
    L_{ce} = \sum_{k \in \{1,2,3 \}} -{y}^k_{ij} \log(\widehat{y}^k_{ij}),
\end{equation}
where ${y}^k_{ij}$ denotes the ground-truth relation category.

\noindent\textbf{Knowledge Distillation.} This step aims to promote the relation prediction capability in a knowledge transfer mechanism. As each relation classifier focuses on distinguishing predicates, especially within the newly added subset, we employ knowledge distillation techniques to preserve the well-learned knowledge from previous stages. During training, we adopt KL-divergence~\cite{kullback1951information} loss to measure the similarity between two relation distributions, and this training objective function is defined as:
{
\setlength\abovedisplayskip{0.5cm}
\setlength\belowdisplayskip{0.5cm}
\begin{equation}
    L_{kl} = \frac{1}{|\mathcal{S}|}\sum_{(k_1,k_2)\in \mathcal{S}} -\widehat{y}^{k_1}_{ij}(\log(\widehat{y}^{k_1}_{ij}) - \log(\widehat{y}^{k_2\prime}_{ij})),
\end{equation}
}
where $\mathcal{S}$ represents the set of pairwise knowledge matching from the relation classifier $\mathtt{Cls}^{\left(k_1\right)}_{rel}$ to $\mathtt{Cls}^{\left(k_2\right)}_{rel}$, with $k_1 \textless k_2$. As shown in Fig.~\ref{fig:pipeline}(b), $\widehat{y}^{k_2\prime}_{ij}$ represents the sliced relation distribution, excluding newly added subsets absent in the previous classification space $\widetilde{\mathcal{R}}^{k_1}$, to ensure equal dimensions. Similar to~\cite{dong2022stacked}, for the knowledge transfer mode, we offer three options: 1) Neighbor: transfer knowledge between adjacent stages. 2) Top-Down: acquire knowledge from all preceding stages. 3) Bi-Direction: facilitate mutual exchange of knowledge among different stages.

\noindent\textbf{Training and Inference.} During training, the total loss includes the cross-entropy loss $L_{ce}$ and KL loss $L_{kl}$:
{
\setlength\abovedisplayskip{0.5cm}
\setlength\belowdisplayskip{0.5cm}
\begin{equation}
\setlength\abovedisplayskip{1pt}
    L_{total} = L_{ce} + \alpha L_{kl},
\setlength\belowdisplayskip{-5pt}
\end{equation}
}
where $\alpha$ is the balancing factor for each loss. In the evaluation stage, we choose the last relation classifier $\mathtt{Cls}^{\left(3\right)}_{rel}$ to obtain the final relation distribution.

\begin{table*}[!t]
  \centering
    \resizebox{0.9\textwidth}{!}{
    \setlength{\tabcolsep}{2.1mm}
    \begin{tabular}{ll||ccc|ccc|c}
\specialrule{0.05em}{0pt}{0pt}     \hline
\multicolumn{2}{c||}{\multirow{2}{*}{Models}} & \multicolumn{7}{c}{PredCls} \\
\cline{3-9}    \multicolumn{2}{c||}{} & \multicolumn{1}{c}{R@20} &\multicolumn{1}{c}{R@50}&\multicolumn{1}{c|}{R@100} & \multicolumn{1}{c}{mR@20}& \multicolumn{1}{c}{mR@50} & \multicolumn{1}{c|}{mR@100} & \multicolumn{1}{c}{Mean} \\
    \hline
    \hline
    \multicolumn{5}{l}{Two-Stage Model-Specific Methods} \\
    \hline
    IMP~\cite{xu2017scene}&{$_{\textit{CVPR'17}}$}   & 31.9 & 36.8 & 38.9 & 9.6 & 10.9 & 11.6 & 23.3 \\
    GPSNet~\cite{lin2020gps}&{$_{\textit{CVPR'20}}$}  & 31.5 & 39.9 & 44.7 & 13.2 & 16.4 & 18.3 & 27.3 \\
    C-SGG~\cite{jin2023fast}&{$_{\textit{CVPR'23}}$}   & 36.5 & - & 46.5 & 32.5 & - & 36.4 & 38.0 \\
\hline
    \multicolumn{5}{l}{Two-Stage Model-Agnostic Methods} \\
    \hline          Motifs~\cite{zellers2018neural}&{$_{\textit{CVPR'18}}$} & \textbf{44.9} & \textbf{50.4} & \textbf{52.4} & 20.2 & 22.1 & 22.9 & 35.5\\
\ \ +BGNN~\cite{li2021bipartite}&{$_{\textit{CVPR'21}}$}& 43.0 & 48.3 & 50.3 & 24.3 & 26.2 & 27.0 & 36.5\\
\ \ +GCL~\cite{dong2022stacked}&{$_{\textit{CVPR'22}}$}& 34.3 & 39.6 & 41.6 & 22.3 & 24.9 & 26.1 & 31.5\\
\ \ +BAI~\cite{DBLP:conf/mm/LiWQJL23}&$_{\textit{MM'23}}$& 39.2 & 44.2 & 46.2 & 31.5 & 35.2 & 36.7 & 38.8 \\
     \rowcolor{mygray-bg} \ \ \textbf{+CAFE (Ours)}& & 39.2 & 45.3 & 47.2 & \textbf{33.2} & \textbf{36.2} & \textbf{36.9} & \textbf{39.7}\\
\hline      VCTree~\cite{tang2019learning}&{$_{\textit{CVPR'19}}$} & \textbf{45.3} & \textbf{50.8} & \textbf{52.7} & 20.5 & 22.6 & 23.3 & 35.9\\
\ \ +BGNN~\cite{li2021bipartite}&{$_{\textit{CVPR'21}}$} & 42.0 & 47.4 & 49.3 & 26.8 & 29.1 & 30.0 & 37.4\\
\ \ +GCL~\cite{dong2022stacked}&{$_{\textit{CVPR'22}}$} & 34.8 & 40.4 & 42.5 & 22.9 & 25.8 & 26.8 & 32.2\\
\ \ +BAI~\cite{DBLP:conf/mm/LiWQJL23}&$_{\textit{MM'23}}$& 39.3 & 44.2 & 46.1 & 30.1 & 35.2 & 36.7 & 38.6 \\
          \rowcolor{mygray-bg} \ \ \textbf{+CAFE (Ours)}& & 40.4 & 46.0 & 48.1 & \textbf{34.6} & \textbf{36.8} & \textbf{37.6} & \textbf{40.6}  \\
\hline      Transformer~\cite{vaswani2017attention}&{$_{\textit{NIPS'17}}$} & 36.4 & 42.2 & 44.5 & 13.2 & 15.2 & 15.9 & 27.9\\
\ \ +BGNN~\cite{li2021bipartite}&{$_{\textit{CVPR'21}}$} & 42.9 & 48.5 & 50.5 & 22.8 & 24.7 & 25.4 & 35.8\\
\ \ +GCL~\cite{dong2022stacked}&{$_{\textit{CVPR'22}}$} & 38.7 & 43.6 & 45.4 & 24.3 & 26.4 & 27.4 & 34.3\\
          \rowcolor{mygray-bg} \ \ \textbf{+CAFE (Ours)}& & \textbf{43.3} & \textbf{48.8} & \textbf{50.9} & \textbf{30.9} & \textbf{32.8} & \textbf{33.5} & \textbf{40.0}  \\
    \specialrule{0.05em}{0pt}{0pt}     \hline
    \end{tabular}%
    }
    \vspace{-0.5em}
    \caption{Performance (\%) of state-of-the-art PSG models on the PSG dataset under the PredCls setting.}
  \label{tab:sota_predcls}%
\end{table*}%

\begin{figure*}[!t]
  \centering
  \includegraphics[width=\linewidth]{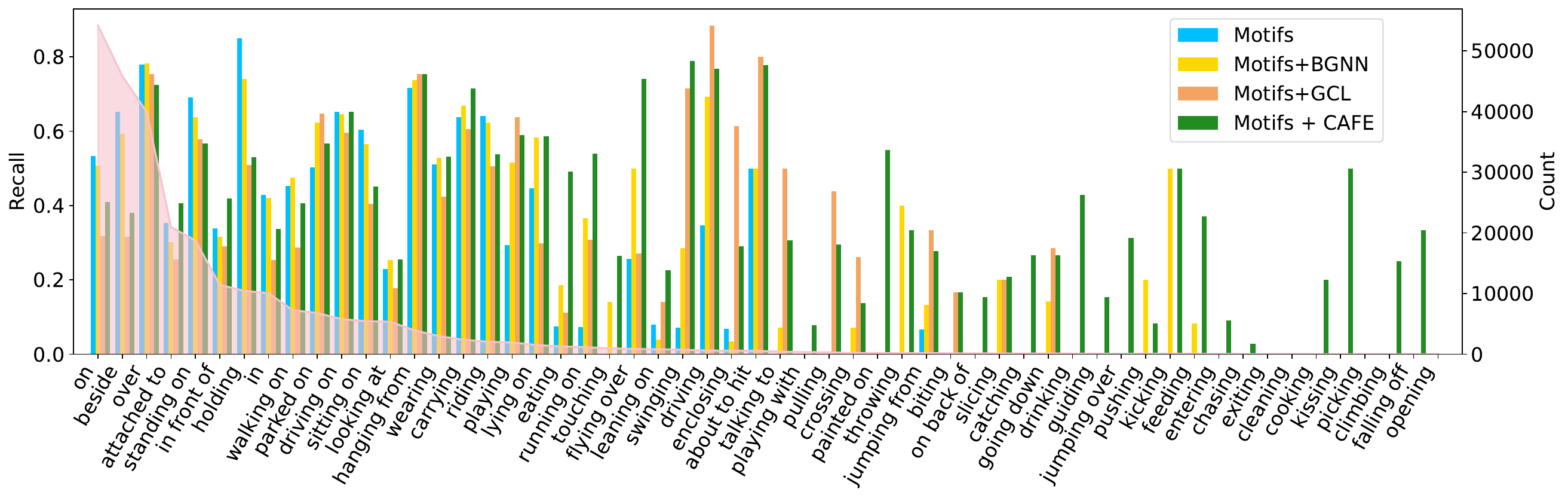}
    \vspace{-2em}
  \caption{Recall statistics for each predicate category on the test set of PSG. The baseline model is Motifs~\cite{zellers2018neural} under PredCls setting. The \textcolor[RGB]{248, 195, 205}{pink} area denotes the predicate distribution of training set.}
  \label{fig:R@100}
\end{figure*}

\begin{table*}[!t]
 \centering
  \resizebox{0.9\textwidth}{!}{
    \setlength{\tabcolsep}{1.5mm}
\begin{tabular}{ll||ccc|ccc|c|c}
\specialrule{0.05em}{0pt}{0pt}     \hline
\multicolumn{2}{c||}{\multirow{2}{*}{Models}} & \multicolumn{8}{c}{SGDet} \\
\cline{3-10}    \multicolumn{2}{c||}{} & \multicolumn{1}{c}{R@20} &\multicolumn{1}{c}{R@50}&\multicolumn{1}{c|}{R@100} & \multicolumn{1}{c}{mR@20}& \multicolumn{1}{c}{mR@50} & \multicolumn{1}{c|}{mR@100} & \multicolumn{1}{c|}{Mean} & \multicolumn{1}{c}{PQ}\\
    \hline
    \hline
    \multicolumn{5}{l}{One-Stage Methods} \\
\hline
 PSGTR~\cite{yang2022panoptic}&{$_{\textit{ECCV'22}}$} &  3.82 & 4.16 & 4.27 & 1.29 & 1.54 & 1.57 & 2.8 & 13.9\\
PSGTR$^\dagger$~\cite{yang2022panoptic}&{$_{\textit{ECCV'22}}$} &  28.4 & 34.4 & 36.3 & 16.6 & 20.8 & 22.1 & 26.4 & 13.9\\
PSGFormer~\cite{yang2022panoptic}&{$_{\textit{ECCV'22}}$} &  16.8 & 19.2 & 20.2 & 14.5 & 17.4 & 18.7 & 17.8 & 36.8\\
PSGFormer$^\dagger$~\cite{yang2022panoptic}&{$_{\textit{ECCV'22}}$} &  18.0 & 19.6 & 20.1 & 14.8 & 17.0 & 17.6 & 17.9 & 36.8\\
\modified{CATQ~\cite{DBLP:conf/mmasia/XuCY23}}&\modified{$_{\textit{MMAsia'23}}$}& \modified{34.8} & \modified{29.7} & \modified{40.3} & \modified{20.9} & \modified{24.9} & \modified{25.2} & \modified{29.3} & \modified{35.9}\\
\modified{Pair-Net$^*$~\cite{DBLP:journals/corr/abs-2307-08699}}&\modified{$_{\textit{arXiv'23}}$}& \modified{24.7} & \modified{28.5} & \modified{30.6} & \modified{29.6} & \modified{35.6} & \modified{39.6} & \modified{31.4} & \modified{40.2}\\
\modified{HiLo$^*$~\cite{zhou2023hilo}}&\modified{$_{\textit{ICCV'23}}$}& \modified{34.1} & \modified{40.7} & \modified{43.0} & \modified{23.7} & \modified{30.3} & \modified{33.1} & \modified{34.2} & \modified{55.4}\\
     \hline
    \multicolumn{5}{l}{Two-Stage Model-Specific Methods} \\
    \hline
    IMP~\cite{xu2017scene}&{$_{\textit{CVPR'17}}$}  & 16.5 & 18.2 & 18.6 & 6.5 & 7.1 & 7.2 & 12.4 & 40.2\\
    GPSNet~\cite{lin2020gps}&{$_{\textit{CVPR'20}}$} & 17.8 & 19.6 & 20.1 & 7.0 & 7.5 & 7.7 & 13.3 & 40.2\\
    C-SGG~\cite{jin2023fast}&{$_{\textit{CVPR'23}}$} & 18.1 & - & 21.6 & 16.6 & - & 17.8 & 18.5 & 40.2\\
   \hline
    \multicolumn{5}{l}{Two-Stage Model-Agnostic Methods} \\
    \hline
        Motifs~\cite{zellers2018neural}&{$_{\textit{CVPR'18}}$}  & 20.0 & 21.7 & 22.0 & 9.1 & 9.6 & 9.7 & 15.3 & 40.2\\
          \ \ +IETrans~\cite{zhang2022fine}&{$_{\textit{ECCV'22}}$}  & 16.7 & 18.3 & 18.8 & 15.3 & 16.5 & 16.7 & 17.1 & 40.2\\
          \ \ +ADTrans~\cite{li2023panoptic}&{$_{\textit{arXiv'23}}$}& 17.1 & 18.6 & 19.0 & 17.1 & 18.0 & 18.5 & 18.1 & 40.2\\
          \ \ \modified{+BAI~\cite{DBLP:conf/mm/LiWQJL23}}&\modified{$_{\textit{MM'23}}$}& \modified{17.4} & \modified{19.0} & \modified{19.4} & \modified{16.7} & \modified{17.6} & \modified{17.7} & \modified{18.0} & \modified{-}\\
          \ \ \modified{+DWIL~\cite{10447193}}&\modified{$_{\textit{ICASSP'24}}$}& \modified{16.5} & \modified{18.0} & \modified{18.2} & \modified{18.4} & \modified{19.0} & \modified{19.2} & \modified{18.2} & \modified{40.2}\\
          
          \rowcolor{mygray-bg}
          \ \ \textbf{+CAFE (Ours)}& &  \textbf{ 23.4} & \textbf{25.7} & \textbf{26.4} & \textbf{25.5} & \textbf{26.5} & \textbf{26.8} & \textbf{25.7}  & \textbf{54.9} \\
    \hline
    VCTree~\cite{tang2019learning}&{$_{\textit{CVPR'19}}$}  & 20.6 & 22.1 & 22.5 & 9.7 & 10.2 & 10.2 & 15.9 & 40.2\\
    \ \ +IETrans~\cite{zhang2022fine}&{$_{\textit{ECCV'22}}$}  & 17.5 & 18.9 & 19.3 & 17.1 & 18.0 & 18.1 & 18.2 & 40.2\\
    \ \ +ADTrans~\cite{li2023panoptic}&{$_{\textit{arXiv'23}}$} & 17.9 & 19.5 & 19.9 & 18.0 & 18.9 & 19.0 & 18.9 & 40.2\\
    \ \ \modified{+RCpsg~\cite{DBLP:conf/mm/YangWLWWYC23}}&\modified{$_{\textit{MM'23}}$}& \modified{22.3} & \modified{24.2} & \modified{24.6} & \modified{11.0} & \modified{11.5} & \modified{11.8} & \modified{17.6} & \modified{40.7}\\
    \ \ \modified{+BAI~\cite{DBLP:conf/mm/LiWQJL23}}&\modified{$_{\textit{MM'23}}$}& \modified{17.9} & \modified{19.5} & \modified{19.9} & \modified{18.0} & \modified{18.9} & \modified{19.0} & \modified{18.9} & \modified{-}\\
    \ \ \modified{+DWIL~\cite{10447193}}&\modified{$_{\textit{ICASSP'24}}$}& \modified{17.7} & \modified{18.1} & \modified{18.3} & \modified{18.3} & \modified{18.9} & \modified{19.0} & \modified{18.4} & \modified{40.2}\\
    \rowcolor{mygray-bg}
    \ \ \textbf{+CAFE (Ours)}& & \textbf{26.9} & \textbf{29.8} & \textbf{30.6}  & \textbf{27.8} & \textbf{29.1} & \textbf{29.4} & \textbf{28.9}  & \textbf{54.9}\\
    Transformer~\cite{vaswani2017attention}&{$_{\textit{NIPS'17}}$} & 19.6 & 21.1 & 21.6 & 8.7 & 9.1 & 9.2 & 14.9 & 40.2\\
    \ \ \modified{+RCpsg~\cite{DBLP:conf/mm/YangWLWWYC23}}&\modified{$_{\textit{MM'23}}$}& \modified{22.7} & \modified{24.3} & \modified{25.0} & \modified{10.9} & \modified{12.0} & \modified{12.4} & \modified{17.9} & \modified{40.8}\\
    \rowcolor{mygray-bg}
    \ \ \textbf{+CAFE (Ours)}& & \textbf{24.6} & \textbf{27.6} & \textbf{28.7}  & \textbf{25.0} & \textbf{26.6} & \textbf{26.9} & \textbf{26.6}  & \textbf{54.9}\\
    \specialrule{0.05em}{0pt}{0pt}     \hline
    \end{tabular}
    }
\vspace{-0.5em}
\caption{Performance (\%) of state-of-the-art PSG models on the PSG dataset under the SGDet setting. Models are trained using 12 epochs by default, while $^\dagger$ denotes the model is trained using 60 epochs. \modified{The detector of one-stage methods is DETR~\cite{DBLP:conf/eccv/CarionMSUKZ20} by default, while $^*$ denotes the detector is Mask2Former~\cite{DBLP:conf/cvpr/ChengMSKG22}\protect\footnotemark.}}
\label{tab:sota_sgdet}%
\end{table*}%

\section{Experiments}\label{sec4}

\subsection{Experimental Settings}
\noindent\textbf{Datasets.} We conducted experiments on the challenging Panoptic Scene Graph Generation (PSG) dataset~\cite{yang2022panoptic}, which contains 48,749 images with 133 object classes (80 thing and 53 stuff classes) and 56 relation classes. Each image is annotated with panoptic segmentation and scene graphs. Our data processing pipelines closely align with~\cite{yang2022panoptic}.

\noindent\textbf{Tasks.} Follow~\cite{yang2022panoptic}, we evaluated the model on two tasks: 1) \emph{Predicate Classification} (\textbf{PredCls}): Given all ground-truth object labels and localizations, we need to predict pairwise predicate categories. 2) \emph{Scene Graph Generation} (\textbf{SGDet}): Given an image, we need to detect all objects and predict both the object categories and their pairwise predicates.

\noindent\textbf{Metrics.} \modified{For robust PSG,} we evaluated the model on three classic metrics: 1) \emph{Recall@K} (\textbf{R@K}): It calculates the proportion of ground-truths that appear among the top-K confident predicted relation triplets. Following prior work, we used $K = \{20, 50, 100\}$. 2) \emph{mean Recall@K} (\textbf{mR@K}): It is the average of R@K scores that are calculated for each predicate category separately, \ie, it puts relatively more emphasis on the tail predicates. 3) \textbf{Mean}: It is the average of all R@K and mR@K scores. Since R@K favors head predicates and mR@K favors tail predicates, Mean is a comprehensive metric that can better evaluate the overall performance~\cite{li2022devil}. \modified{For zero-shot PSG, we adopted two metrics: 1) \textit{Zero-Shot Recall@K} (\textbf{zR@K}): It only calculates the R@K for subject-predicate-object triplets that have not occurred in the training set, offering a focused measure of the generalization ability to novel instances~\cite{lu2016visual}. 2) \textbf{Average}: It calculates the average value of all zR@K scores, which is a comprehensive metric to assess the zero-shot learning ability.} In the SGDet task, we use panoptic segmentation evaluation protocol \textbf{PQ}~\cite{kirillov2019panoptic} as an auxiliary metric.

\subsection{Implementation Details} 
\noindent \textbf{Mask Generation Module.} We adopted Panoptic FPN~\cite{kirillov2019panoptic} as our panoptic segmentor in the PredCls task and adopted Panoptic Segformer~\cite{li2022panoptic} pretrained on COCO~\cite{lin2014microsoft} in the SGDet task. Our choice for the backbone was ResNet-50-FPN~\cite{he2016deep}, and RoI features were extracted using the RoIAlign operation~\cite{he2017mask}. We kept the parameters of the panoptic segmentor frozen during training to avoid potential biases introduced by the training data. 

\noindent \textbf{Object Classification Module.} Our context module aligned with the baseline (\eg, Motifs~\cite{zellers2018neural}, using bi-directional LSTM structures). This classification module was trained on the standard cross-entropy loss given the object class labels. 

\noindent \textbf{Relation Classification Module.} Our relation classification module employed the CAFE framework, which encompassed the shape-aware feature preparation and curricular feature learning. Regarding hyperparameters, we set $\alpha$ to 1.0, $\lambda$ to 0.5, and $\mu$ to 0.8. As for training, we utilized the SGD optimizer with a mini-batch size of 16 and an initial learning rate of 0.002, conducting training for 12 epochs. We employed a simple concatenation feature fusion strategy for the Motifs~\cite{zellers2018neural} and VCTree~\cite{tang2019learning} baselines, while adopting an entangled attention approach under the Transformer~\cite{vaswani2017attention}. We adopted the Top-Down knowledge transfer mode and the same warm-up and decayed strategy as~\cite{yang2022panoptic}. All experiments were conducted with NVIDIA 2080ti GPUs.

\subsection{Comparison with State-of-the-Arts in Robust PSG}

\begin{table}[!t]
  \begin{center}
  \resizebox{0.48\textwidth}{!}{
    \setlength{\tabcolsep}{2.7mm}
    \begin{tabular}{l||c|c|c|c}
    \specialrule{0.05em}{0pt}{0pt}
    \hline
    \multicolumn{1}{c||}{\multirow{2}{*}{Models}} & \multicolumn{4}{c}{PredCls} \\
\cline{2-5}          & Head  & Body  & Tail  & Avg \\
    \hline
    \hline
    Motifs~\cite{zellers2018neural} & \textbf{56.3}  & 13.1  & 0.3   & 23.2  \\
    \ \ +BGNN~\cite{li2021bipartite} & 55.4  & 25.8  & 5.7   & 29.0  \\
    \ \ \textbf{+CAFE}   & 50.9  & \textbf{45.7}  & \textbf{16.8}  & \textbf{37.8}  \\
    \specialrule{0.05em}{0pt}{0pt}
    \hline
    \end{tabular}
    }
    \end{center}
    \vspace{-1.5em}
    \caption{Recall@100 of each predicate group under PredCls setting. Avg is the average of three groups.}
  \label{tab:R@100}
\end{table}%

\begin{figure*}[!t]
  \centering
  \includegraphics[width=\linewidth]{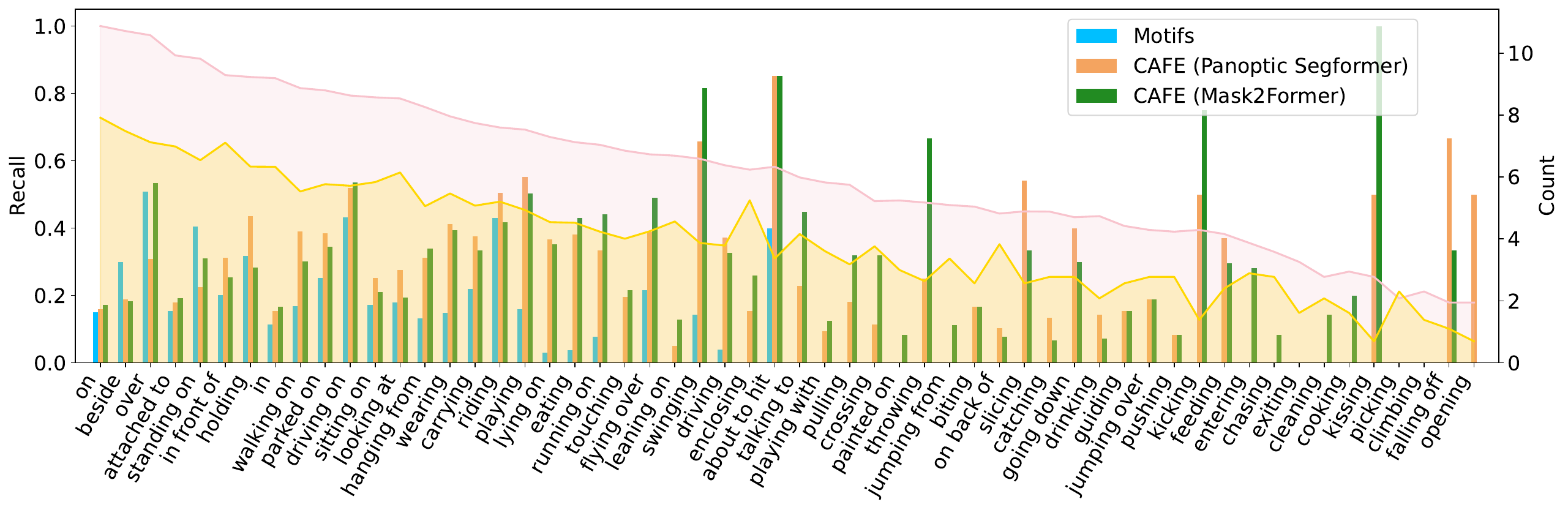}
    \vspace{-2em}
  \caption{\modified{Recall statistics for each predicate category on the test set of PSG. The baseline model is Motifs under SGDet setting. The \textcolor[RGB]{248, 195, 205}{pink} area denotes the predicate distribution of training set, while the \textcolor[RGB]{255, 215, 0}{yellow} area denotes the predicate distribution of test set. 
For better visualization, the logarithm of the predicate distribution is taken for both the training set and the test set.} }
  \label{fig:sgdet_2}
\end{figure*}

\noindent\textbf{Setting.} Due to the model-agnostic nature, we equipped our CAFE with three popular baselines: \textbf{Motifs}~\cite{zellers2018neural}, \textbf{VCTree}~\cite{tang2019learning} and \textbf{Transformer}~\cite{vaswani2017attention}. We compared our methods with various state-of-the-art PSG models on the PSG dataset under PredCls setting (Table~\ref{tab:sota_predcls}) and SGDet setting (Table~\ref{tab:sota_sgdet}). Specifically, these models can be divided into three groups: 1) One-stage methods: \textbf{PSGTR}~\cite{yang2022panoptic}, \textbf{PSGFormer}~\cite{yang2022panoptic}, \modified{\textbf{CATQ}~\cite{jin2023fast}, \textbf{Pair-Net}~\cite{DBLP:journals/corr/abs-2307-08699} and \textbf{HiLo}~\cite{zhou2023hilo}}. 2) Two-stage model-specific methods: \textbf{IMP}~\cite{xu2017scene}, \textbf{GPSNet}~\cite{lin2020gps} and \textbf{C-SGG}~\cite{jin2023fast}. 3) Two-stage model-agnostic methods: \textbf{BGNN}~\cite{li2021bipartite}, \textbf{GCL}~\cite{dong2022stacked}, \textbf{IETrans}~\cite{zhang2022fine}, \textbf{ADTrans}~\cite{li2023panoptic}, \modified{\textbf{BAI}~\cite{DBLP:conf/mm/LiWQJL23}, \textbf{DWIL}~\cite{10447193} and \textbf{RCpsg}~\cite{DBLP:conf/mm/YangWLWWYC23}}. \modified{One-stage methods generally achieve better performance under the SGDet setting due to their trainable object detectors and end-to-end training schemes, but consume more computational resources (\eg, GPU and training time).} \footnotetext{\modified{The performance improvement may stem from the superior panoptic segmentation mask generation capability of Mask2Former, rather than solely from the model itself.}} For a fair comparison, we compared with the methods in the last two groups. To further illustrate the performance of our model, we presented recall statistics for each predicate of CAFE under the PredCls setting, alongside the baseline~\cite{zellers2018neural} and BGNN~\cite{li2021bipartite} in Fig.~\ref{fig:R@100} for comparative analysis. \modified{We also presented the predicate distributions for both the training and test sets in Fig.~\ref{fig:sgdet_2} using a logarithmic transformation for better visual clarity, and detail the recall statistics under the SGDet setting.} Besides, we reported the Recall@100 metric of each predicate category group\footnote{We divided predicates into Head, Body, and Tail groups based on the distribution in different GT annotations in PSG dataset.} under PredCls setting in Table~\ref{tab:R@100}.

\noindent\textbf{Quantitative Results in PredCls.} From the results under the PredCls setting in Table~\ref{tab:sota_predcls}, we find that: 1) The two strong baselines (\ie, Motifs and VCTree) achieve the best R@K due to frequency bias. Nonetheless, as shown in Table~\ref{tab:R@100}, these baselines suffer severe drops in tail predicates (\eg, 0.3\% in Tail \vs~23.2\% in Avg on the R@100 metric). 2) Due to the common label noises in dataset (\eg, head predicate \texttt{on} and tail predicate \texttt{lying on} are all reasonable for \texttt{man}-\texttt{bed}, but the only GT is \texttt{on}), the improvement of mR@K will inevitably drop the scores in R@K. Despite this minor decrease in R@K, CAFE exhibits a substantial improvement on mR@K metric compared to the baselines and outperforms the SOTA model-specific methods (\eg, C-SGG). 3) CAFE achieves a better trade-off between the R@K and mR@K, surpassing the SOTA methods in the Mean metric (\eg, 38.0\% in C-SGG \vs~40.6\% in CAFE based on VCTree). Additionally, as shown in Table~\ref{tab:R@100} and Fig.~\ref{fig:R@100}, CAFE exhibits slight sacrifice on head predicates but achieves notable improvement for tail predicates, demonstrating overall superiority across all predicates. 

\noindent\textbf{Quantitative Results in SGDet.} From the results under the SGDet setting in Table~\ref{tab:sota_sgdet}, we can observe that: 1) Our CAFE \modified{outperforms the latest two-stage SOTA methods across all scene graph evaluation metrics, achieving the best trade-off between R@K and mR@K metrics (\eg, 18.4\% in DWIL \vs~28.9\% in CAFE based on VCTree, 17.9\% in RCpsg \vs~26.6\% in CAFE based on Transformer). 2) CAFE can surpass nearly all leading one-stage methods in terms of the PQ metric (\eg, 36.8\% in PSGFormer \vs~54.9\% in CAFE). Furthermore, as depicted in Fig.~\ref{fig:sgdet_2}, the predicate distributions of the training and test sets clearly exhibit a distributional shift that our model must adapt to. Such shifts are characteristic challenges in Robust Learning, akin to the adversarial perturbations and corruptions that models may encounter in real-world scenarios. Even in these conditions, CAFE is capable of predicting the vast majority of predicates and exceeds baseline performance in over 95\% of the categories, underscoring its effectiveness and robustness.}
\begin{figure}[!t]
  \centering
  \includegraphics[width=\linewidth]{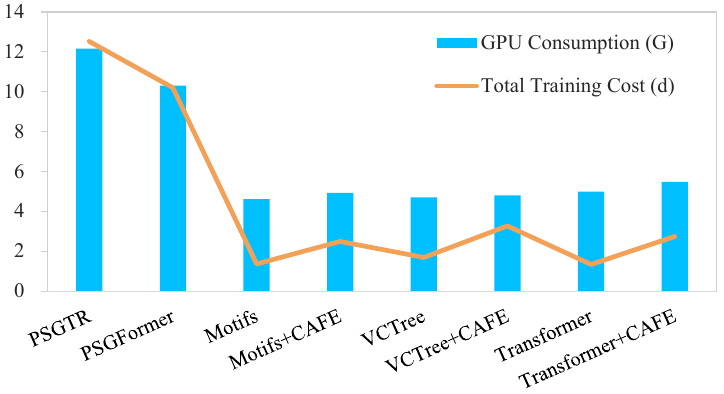}
  \vspace{-2em}
  \caption{\modified{The Statistics of GPU consumption and total training cost, under the SGDet setting.}} 
  \label{fig:cost}
\end{figure}

\begin{table*}[!t]
  \begin{center}
  \resizebox{0.9\textwidth}{!}{
    \setlength{\tabcolsep}{3mm}
    \begin{tabular}{l||cc|cc|cc|cc}
    \specialrule{0.05em}{0pt}{0pt}
    \hline
    \multicolumn{1}{c||}{\multirow{2}{*}{Models}} & \multicolumn{8}{c}{PredCls}\\
\cline{2-9}          & zR@20  & \textcolor[rgb]{ .753,  0,  0}{$\scriptstyle \bigtriangleup$} & zR@50  & \textcolor[rgb]{ .753,  0,  0}{$\scriptstyle \bigtriangleup$}& zR@100 & \textcolor[rgb]{ .753,  0,  0}{$\scriptstyle \bigtriangleup$}  & Average & \textcolor[rgb]{ .753,  0,  0}{$\scriptstyle \bigtriangleup$} \\
    \hline
    \hline
    Motifs~\cite{zellers2018neural} &25.6 & \textcolor[rgb]{ .753,  0,  0}{-}  &37.2 & \textcolor[rgb]{ .753,  0,  0}{-} & 42.3 & \textcolor[rgb]{ .753,  0,  0}{-}&  35.0 & \textcolor[rgb]{ .753,  0,  0}{-}\\
    \ \ +CAFE   & \textbf{31.4} & \textcolor[rgb]{ .753,  0,  0}{5.8} & \textbf{42.2}& \textcolor[rgb]{ .753,  0,  0}{5.0} & \textbf{44.1} & \textcolor[rgb]{ .753,  0,  0}{1.8}&  \textbf{39.2} & \textcolor[rgb]{ .753,  0,  0}{4.2}  \\
    \hline
    VCTree~\cite{tang2019learning} & 28.2 & \textcolor[rgb]{ .753,  0,  0}{-} & 33.3 & \textcolor[rgb]{ .753,  0,  0}{-}& 39.7& \textcolor[rgb]{ .753,  0,  0}{-} & 33.8 & \textcolor[rgb]{ .753,  0,  0}{-} \\
    \ \ +CAFE   & \textbf{30.4} & \textcolor[rgb]{ .753,  0,  0}{2.2} & \textbf{40.2} & \textcolor[rgb]{ .753,  0,  0}{6.9}& \textbf{42.2} & \textcolor[rgb]{ .753,  0,  0}{2.5}&  \textbf{37.6} & \textcolor[rgb]{ .753,  0,  0}{3.8}  \\
    \hline
    Transformer~\cite{vaswani2017attention} & 20.5& \textcolor[rgb]{ .753,  0,  0}{-}  & 42.3& \textcolor[rgb]{ .753,  0,  0}{-} &  50.0& \textcolor[rgb]{ .753,  0,  0}{-}&  37.6 & \textcolor[rgb]{ .753,  0,  0}{-}\\
    \ \ +CAFE   &  \textbf{38.9}& \textcolor[rgb]{ .753,  0,  0}{18.4} & \textbf{47.7}& \textcolor[rgb]{ .753,  0,  0}{5.4} & \textbf{51.6} & \textcolor[rgb]{ .753,  0,  0}{1.6}&   \textbf{46.1}  & \textcolor[rgb]{ .753,  0,  0}{8.5} \\
    \specialrule{0.05em}{0pt}{0pt}
    \hline
    \end{tabular}
    }
    \end{center}
    \vspace{-1.5em}
    \caption{\modified{Performance (\%) of CAFE in zero-shot PSG scenarios under the PredCls setting.}}
  \label{tab:zR@K_predcls}
\end{table*}%

\noindent\textbf{Statistics of Computational Cost.} \modified{To provide an intuitive representation of the differences in GPU consumption (G) and total training cost (d) among different PSG models (\ie, both one-stage and two-stage models), we visualized the results under the SGDet setting in Fig.~\ref{fig:cost}. The above results were obtained using a single 3090Ti GPU in a PyTorch 1.9 and CUDA 11.0 environment, with the batch size set to 1. As shown in Fig.~\ref{fig:cost}, we can observe that: 1) Compared to one-stage PSG methods (\eg, PSGTR and PSGFormer), our CAFE can save over 50\% of GPU memory resources and reduce total training time by over 65\%. Additionally, these one-stage PSG methods require training for 60 epochs, while our CAFE achieves convergence in just 12 epochs or even fewer, and also surpasses them in terms of performance. 2) Compared to two-stage baselines, while our CAFE incurs a slight increase in GPU consumption and total training time, it demonstrates superior model performance over the baseline models.}

\begin{figure}[!t]
  \centering
  \includegraphics[width=\linewidth]{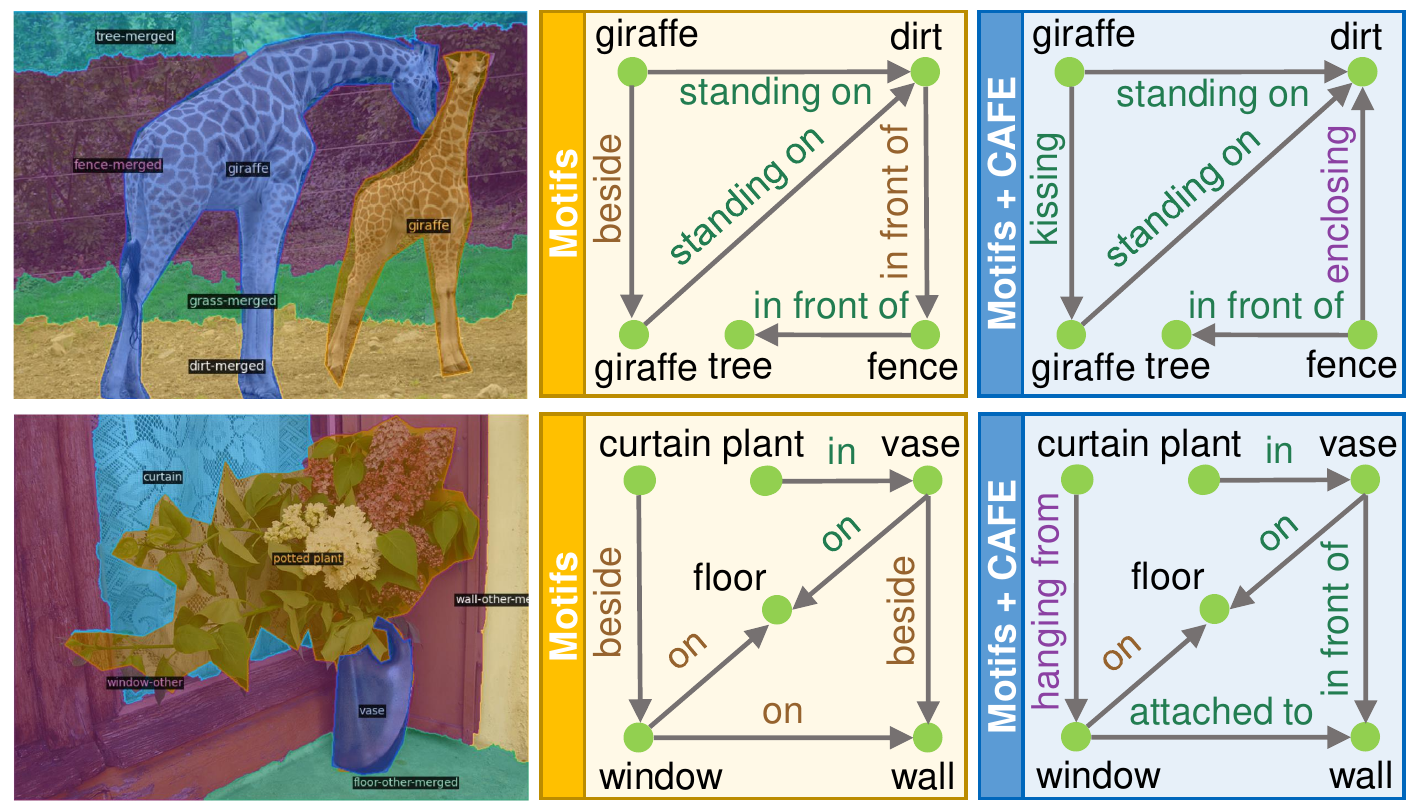}
    \vspace{-1.5em}
  \caption{The visualization results of scene graphs generated by Motifs (yellow) and Motifs+CAFE (blue). \textcolor[RGB]{34,125,81}{Green} predicates are correct (\ie, match GT), \textcolor[RGB]{150,99,46}{brown} predicates are acceptable (\ie, does not match GT but still reasonable), and \textcolor[RGB]{139,64,162}{purple} predicates are more informative (\ie, not the same as GT but more informative).}
  \label{fig:keshihua}
\end{figure}

\noindent\textbf{Qualitative Analysis.} We visualized some qualitative results generated by Motifs and Motifs+CAFE under PredCls setting in Fig.~\ref{fig:keshihua}. The results give the following insights: 1) Motifs-baselines are biased towards coarse-grained predicates (\eg, \texttt{$\langle$giraffe-beside-giraffe$\rangle$}) caused by long-tailed distribution. However, CAFE can make accurate predictions (\eg, \texttt{$\langle$giraffe-kissing-giraffe$\rangle$}) by leveraging shape-aware features that capture interactions between object pairs. 2) CAFE can provide more fine-grained and informative predicates due to the curriculum learning strategy that learns from easy to hard (\eg, \texttt{hanging from} \vs~\texttt{beside}, and \texttt{enclosing} \vs~\texttt{in front of}).

\begin{table*}[!t]
  \centering
  \resizebox{0.9\textwidth}{!}{
    \setlength{\tabcolsep}{2.2mm}
    \begin{tabular}{ccc||ccc|ccc|c}
     \specialrule{0.05em}{0pt}{0pt}     \hline
    \multicolumn{3}{c||}{Feature} & \multicolumn{7}{c}{PredCls} \\
    \hline
    
    bbox&mask&boundary& \multicolumn{1}{c}{R@20} &\multicolumn{1}{c}{R@50}&\multicolumn{1}{c|}{R@100} & \multicolumn{1}{c}{mR@20}& \multicolumn{1}{c}{mR@50} & \multicolumn{1}{c|}{mR@100} & \multicolumn{1}{c}{Mean} \\
    \hline\hline
    \Checkmark& & & 37.2 & 42.9 & 45.0  & 28.2 & 30.5 & 31.5  & 35.9  \\
     &\Checkmark& & 38.5 & 44.4 & 46.8  & 29.7 & 32.1 & 33.0  & 37.4  \\
     & &\Checkmark& 39.1 & 44.7 & 46.7  & 29.8 & 31.8 & 32.5  & 37.5  \\
    \Checkmark&\Checkmark& & 38.9 & 44.8 & 47.0  & 30.4 & 33.3 & 34.3  & 38.1  \\
    \Checkmark& &\Checkmark& 39.4 & 45.0 & 47.1  & 31.0 & 33.0 & 33.9  & 38.3  \\
    &\Checkmark&\Checkmark& \textbf{39.7} & \textbf{45.4} & \textbf{47.5}  & 31.1 & 33.4 & 34.0  & 38.5  \\
    \Checkmark&\Checkmark&\Checkmark& 39.2 & 45.3 & 47.2 & \textbf{33.2} & \textbf{36.2} & \textbf{36.9} & \textbf{39.7}  \\
     \specialrule{0.05em}{0pt}{0pt}     \hline
    \end{tabular}%
    }
    \vspace{-0.5em}
    \caption{Ablation studies on different features of CAFE. The baseline model is Motifs~\cite{zellers2018neural} under PredCls setting.}
  \label{ablation:features}%
\end{table*}%
\begin{table*}[!t]
  \centering
  \resizebox{0.9\textwidth}{!}{
    \setlength{\tabcolsep}{0.6mm}
    \begin{tabular}{l||ccc|ccc|c|cc}
     \specialrule{0.05em}{0pt}{0pt}     \hline
    \multicolumn{1}{c||}{\multirow{2}{*}{Feature Fusion}} & \multicolumn{9}{c}{PredCls} \\
\cline{2-10}          & \multicolumn{1}{c}{R@20} &\multicolumn{1}{c}{R@50}&\multicolumn{1}{c|}{R@100} & \multicolumn{1}{c}{mR@20}& \multicolumn{1}{c}{mR@50} & \multicolumn{1}{c|}{mR@100} & Mean & Training Cost & Testing Cost\\
    \hline
    \hline
    Concatenation  & 37.9 & 43.2 & 45.0 & 27.0 & 28.7 & 29.2    & 35.2 & \textbf{1.95} & \textbf{0.093}\\
    Joint fusion  & 38.2 & 43.5 & 45.4 & 27.2 & 28.8 & 29.3 & 35.4 & 2.07 & 0.093\\
    Divided fusion & 42.8 & 48.0& 49.9  & 27.2 & 27.8 & 29.3  & 37.5 & 2.18 & 0.201\\
    Entangled attention & \textbf{43.3} & \textbf{48.8} & \textbf{50.9}  & \textbf{30.9} & \textbf{32.8} & \textbf{33.5}  & \textbf{40.0} & 2.38 & 0.204\\
     \specialrule{0.05em}{0pt}{0pt}     \hline
    \end{tabular}%
    }
    \vspace{-0.5em}
    \caption{Ablation studies on feature fusion strategies. The baseline model is Transformer~\cite{vaswani2017attention} under PredCls setting.}
   \label{feature}%
\end{table*}%
\subsection{Comparison in Zero-Shot PSG}

\modified{\noindent\textbf{Setting.} In the PSG dataset, the test set contains a total of 874 visual triplets that never occur within the training set. Following~\cite{tang2020unbiased}, we selected these unseen relation triplets of the test set as the evaluated samples. Then, we equipped our CAFE with three popular baselines: \textbf{Motifs}~\cite{zellers2018neural}, \textbf{VCTree}~\cite{tang2019learning} and \textbf{Transformer}~\cite{vaswani2017attention}, to individually assess each model on the task of zero-shot PSG. We reported the comparison results for detecting unseen visual triplets in Table~\ref{tab:zR@K_predcls} under the PredCls setting.

\noindent\textbf{Results.} As the results shown in Table~\ref{tab:zR@K_predcls}, we can observe that our CAFE model exhibits superior performance across all zero-shot evaluation metrics (\eg, 46.1\% in CAFE \vs~37.6\% in Transformer on the Average metric), demonstrating strong generalizability. This performance gap stems from two aspects: 1) Baseline models (\eg, Motifs) heavily rely on statistical prior knowledge regarding the co-occurrence of subject and object categories~\cite{he2022state}. Consequently, they tend to predict high-frequency seen triplets whose relations are conceptually simple (\eg, spatial relations like ``\texttt{over}") due to the skewed predicate distribution. 2) 
CAFE can learn more robust visual relationship features by adaptively integrating different shape-aware features during training. These learned visual relation features between different subject-object pairs exhibit significant variations, thus alleviating the bias on statistics prior. For example, \texttt{over} signifies the subject being positioned above the object with little to no interaction between them, whereas \texttt{enclosing} describes a partial surrounding boundary. It is the precise capturing of object shapes and interactions between objects that enables CAFE to accurately infer unseen visual relation triplets (\eg, \texttt{tree}-\texttt{enclosing}-\texttt{zebra}).
}\\

\subsection{Ablation Studies}
\noindent\textbf{Effectiveness of Shape-Aware Features.} We evaluated the significance of our proposed shape-aware features in the CAFE framework based on Motifs~\cite{zellers2018neural} under the PredCls setting. This framework comprises a training process with three stages, each corresponding to a distinct set of features: represent spatial information (\textbf{bbox}), convey object shapes (\textbf{mask}), and depict interactions of object pairs (\textbf{boundary}). As reported in Table~\ref{ablation:features}, we have the following observations: 1)  Using solely the bbox feature yields the weakest performance, since it ignores the shapes of objects and their interactions. These shape-aware features provide vital guidance for accurate relation predictions. However, bbox features remain essential as they convey spatial information, allowing accurate predictions of certain simple positional predicates (\cf, Sec.~\ref{sec1}). 2) Incrementally increasing the complexity of features contributes to a gradual enhancement of the overall model performance, as indicated by improvements in the Mean metric. 3) While mask features are a superior choice over bbox features for single-feature use (\#1 \vs~\#2), achieving optimal performance requires the combination of features (\#4). The amalgamation of all features yields the best mR@K and the highest Mean score.

\noindent\textbf{Different Feature Fusion Strategies in CAFE.}
We incorporated CAFE into the Transformer baseline~\cite{vaswani2017attention} under the PredCls setting, employing four feature fusion strategies: simple concatenation, joint fusion, divided fusion, and entangled attention \modified{(\cf, Sec.~\ref{3.2.2})}. Additionally, we reported the expenses associated with different feature fusion strategies during the training and inference phases for each image. As in Table~\ref{feature}, we can observe that: 1) Simple concatenation incurs the lowest training cost (1.95s) and inference cost (0.093s). 2) CAFE is robust and effective for various feature fusions. The entangled attention strategy yields the best performance in both mR@K and Mean metrics.

\begin{table}[!t]
  \centering
  \resizebox{0.48\textwidth}{!}{
    \setlength{\tabcolsep}{0.5mm}
    \begin{tabular}{cc||c|c|c}
     \specialrule{0.05em}{0pt}{0pt}     \hline
    \multicolumn{1}{c}{\multirow{2}{*}{CL}} & \multicolumn{1}{c||}{\multirow{2}{*}{Feature}} & \multicolumn{3}{c}{PredCls} \\
\cline{3-5}         & & R@20/50/100 & mR@20/50/100 & Mean \\
    \hline
    \hline
    \XSolidBrush& bbox & 38.6 / 44.1 / 46.3 & 25.6 / 28.1 / 29.0 & 35.3  \\
    \XSolidBrush& mask & 38.9 / 44.9 / 47.0  & 26.4 / 28.8 / 29.5 & 35.9  \\
    \XSolidBrush& boundary& 39.1 / 45.1 / 47.0  & 27.3 / 29.5 / 30.4  & 36.5  \\
    \Checkmark & all & \textbf{39.2} / \textbf{45.3} / \textbf{47.2} & \textbf{33.2} / \textbf{36.2} / \textbf{36.9} & \textbf{39.7}  \\
     \specialrule{0.05em}{0pt}{0pt}     \hline
    \end{tabular}%
    }
    \vspace{-0.5em}
    \caption{\modified{Ablation studies on curriculum learning of CAFE.}}
   \label{ablation: cl}%
\end{table}%
\modified{\noindent\textbf{Effectiveness of Curriculum Learning.} We evaluated the significance of curriculum learning in CAFE based on Motifs~\cite{zellers2018neural} under the PredCls setting. Specifically, during the training phase, we no longer divided the training process into three stages. Instead, we retained only one relation classifier and utilized only one type of feature. Additionally, we exclusively used the cross-entropy loss function, that is, the training objective function is $L_{ce}$ (cf., Eq.~(\textcolor{red}{16})). As the results in Table~\ref{ablation: cl}, we have the following observations: 1) Our proposed curriculum learning strategy yields slight improvements in R@K (\eg, 0.2\% $\sim$ 0.9\% gains in R@100) and significant enhancements in mR@K (\eg, 6.5\% $\sim$ 7.9\% gains in mR@100), as well as notable improvements in the Mean metric (\eg, 3.2\% $\sim$ 4.6\% gains). 2) In scenarios where the curriculum learning strategy is not employed, shape-aware features are a superior choice over bbox features for single-feature use (\eg, 36.5\% \vs~35.3\% in the Mean metric).}

\noindent\textbf{Different Predicate Grouping Modes in CAFE.} We investigated the impact of different predicate grouping modes within the CAFE framework under the PredCls setting with baseline model Motifs~\cite{zellers2018neural}. We considered three predicate grouping modes: 1) Random: random division of predicates into three groups. 2) Average: equal division based on predicate distribution. 3) Cognition-based: initial grouping by predicate distribution, followed by adjustments using semantic similarity. As shown in Table~\ref{ablation: predicate}, we can observe: 1) Average group excels in R@K due to frequency bias, but suffers severe drops in tail predicates. This phenomenon is attributed to the inherent defects in the PSG dataset, making it challenging to achieve better results in both mR@K and R@K. 2) Cognition-based grouping mode exhibits its effectiveness by achieving the highest results in both mR@K and Mean metrics. This is attributed to the allocation of semantically similar predicates to distinct groups, ensuring that each relation classifier excels in distinguishing the predicates within its designated group.

\begin{table}[!t]
  \centering
  \resizebox{0.48\textwidth}{!}{
    \setlength{\tabcolsep}{0.2mm}
    \begin{tabular}{l||c|c|c}
     \specialrule{0.05em}{0pt}{0pt}     \hline
    \multicolumn{1}{c||}{\multirow{2}{*}{Group Mode}} & \multicolumn{3}{c}{PredCls} \\
\cline{2-4}          & R@20/50/100 & mR@20/50/100 & Mean \\
    \hline
    \hline
    Random & 39.8 / 45.4 / 47.3  & 29.2 / 31.8 / 32.4  & 37.7  \\
    Average & \textbf{40.2} / \textbf{45.8} / \textbf{47.9}  & 30.9 / 32.8 / 33.7  & 38.5  \\
    Cognition-based & 39.2 / 45.3 / 47.2 & \textbf{33.2} / \textbf{36.2} / \textbf{36.9} & \textbf{39.7}  \\
     \specialrule{0.05em}{0pt}{0pt}     \hline
    \end{tabular}%
    }
    \vspace{-0.5em}
    \caption{Ablation studies on predicate grouping modes.}
   \label{ablation: predicate}%
\end{table}%
\begin{table}[!t]
  \centering
  \resizebox{0.48\textwidth}{!}{
    \setlength{\tabcolsep}{0.6mm}
    \begin{tabular}{cc||c|c|c}
     \specialrule{0.05em}{0pt}{0pt}     \hline
    \multicolumn{2}{c||}{Sampling} & \multicolumn{3}{c}{PredCls} \\
\cline{1-5}   Over &   Median   & R@20/50/100 & mR@20/50/100 & Mean \\
    \hline
    \hline
     \XSolidBrush&\XSolidBrush &\textbf{42.2} / \textbf{47.8} / \textbf{49.7} & 26.0 / 28.1 / 28.9 & 37.1  \\
    \Checkmark & \XSolidBrush&38.9 / 44.7 / 46.7 & 31.8 / 33.9 / 34.6 & 38.4  \\
    \XSolidBrush & \Checkmark &41.7 / 47.1 / 49.2 & 28.6 / 30.4 / 31.2 & 38.0  \\
    \Checkmark & \Checkmark & 39.2 / 45.3 / 47.2 & \textbf{33.2} / \textbf{36.2} / \textbf{36.9} & \textbf{39.7}  \\
     \specialrule{0.05em}{0pt}{0pt}     \hline
    \end{tabular}%
    }
    \vspace{-0.5em}
    \caption{\modified{Ablation studies on predicate sampling strategies.}}
    \vspace{-0.5em}
   \label{ablation: resampling}%
\end{table}%
\modified{\noindent\textbf{Effectiveness of Predicate Sampling Strategy.} We propose a two-fold predicate sampling strategy to obtain a relatively balanced training set: 1) Oversampling on images, and 2) Median resampling on triplets (\cf, Sec.~\ref{3.3.3}). We conducted ablation studies on these strategies within the CAFE based on Motifs~\cite{zellers2018neural} under PredCls. As reported in Table~\ref{ablation: resampling}, we can observe that: 1) Without any resampling strategy, the model tends to bias towards head predicates (\eg, highest R@K), but shows a notable drop in performance on tail predicates (\eg, lowest mR@K). 2) Both oversampling and median sampling can enhance mR@K (\eg, 2.3\% $\sim$ 5.7\% gains in mR@100) and keep competitive R@K (\eg, 0.5\% $\sim$ 3.0\% loss in R@100). These strategies can alleviate data imbalance to some extent, enhancing the Mean metric (\eg, 0.9\% $\sim$ 1.3\% gains). 3) Combining these two resampling strategies allows for the best trade-off (\eg, highest Mean), confirming that balanced data is crucial for preventing biased predictions and enhancing overall performance.}

\begin{table}[!t]
  \centering
    \resizebox{0.48\textwidth}{!}{
    \setlength{\tabcolsep}{2mm}
    \begin{tabular}{c||c|c|c}
     \specialrule{0.05em}{0pt}{0pt}     \hline
    \multicolumn{1}{c||}{\multirow{2}{*}{$\lambda$}} & \multicolumn{3}{c}{PredCls} \\
\cline{2-4}          & R@20/50/100 & mR@20/50/100 & Mean \\
    \hline
    \hline
    0 & 28.6 / 34.1 / 36.2 & \textbf{36.2} / \textbf{39.2} / \textbf{40.5} & 35.8  \\
    0.25 & 36.9 / 42.3 / 44.6 & 35.0 / 37.0 / 37.8 & 38.9  \\
    0.5 & 39.2 / 45.3 / 47.2 & 33.2 / 36.2 / 36.9 & \textbf{39.7}  \\
    0.75 & 39.3 / 45.5 / 47.4 & 32.5 / 35.2 / 35.9 & 39.3  \\
    1.0 & \textbf{40.0} / \textbf{45.5} / \textbf{47.6} & 31.2 / 33.9 / 34.8 & 38.8  \\
    \specialrule{0.05em}{0pt}{0pt}     \hline
    \end{tabular}%
    }
    \vspace{-0.5em}
    \caption{Ablation studies on hyperparameter $\lambda$.}
    \label{tab:lambda}%
\end{table}%

\begin{table}[!t]
  \centering
    \resizebox{0.48\textwidth}{!}{
    \setlength{\tabcolsep}{0.4mm}
    \begin{tabular}{l||c|c|c}
     \specialrule{0.05em}{0pt}{0pt}     \hline
    \multicolumn{1}{c||}{\multirow{2}{*}{Transfer Mode}} & \multicolumn{3}{c}{PredCls} \\
\cline{2-4}          & R@20/50/100 & mR@20/50/100 & Mean \\
    \hline
    \hline
    Neighbor & 38.2 / 44.1 / 45.9 & 32.1 / 34.5 / 35.1 & 38.3  \\
    Top-Down & \textbf{39.2} / \textbf{45.3} / \textbf{47.2} & 33.2 / 36.2 / 36.9 & \textbf{39.7}  \\
    Bi-Direction & 38.0 / 43.8 / 45.8 & \textbf{34.1} / \textbf{36.4} / \textbf{37.2} & 39.2  \\
    \specialrule{0.05em}{0pt}{0pt}     \hline
    \end{tabular}%
    }
    \vspace{-0.5em}
    \caption{Ablation studies on knowledge transfer modes.}
    \label{tab:knowledge}%
\end{table}%

\noindent\textbf{Hyperparameter $\lambda$ in Predicate Sampling.} As mentioned in Sec.~\ref{3.3.3}, $\lambda$ serves as a balancing factor to regulate the sampling rates between $\phi_1^k$ and $\phi_2^k$. We investigated $\lambda \in \{0, 0.25, 0.5, 0.75, 1.0\}$ under the PredCls setting with Motifs~\cite{zellers2018neural} in Table~\ref{tab:lambda}. From the results in Table~\ref{tab:lambda}, it is evident that when $\lambda$ is excessively small, the model inclines towards tail predicates, resulting in a high mR@K. Conversely, when $\lambda$ is excessively large, the model is biased towards head predicates, yielding a high R@K. To better trade-off the performance on different predicates (\ie, the highest Mean), we set $\lambda$ to 0.5.

\noindent\textbf{Different Knowledge Transfer Modes in CAFE.} We conducted three modes under the PredCls setting with Motifs~\cite{zellers2018neural} to assess the effects of knowledge transfer methods, as presented in Table~\ref{tab:knowledge}. We can observe that the Bi-Direction mode achieves the best mR@K due to its capacity for mutual exchange and knowledge influence across different stages. For a balanced performance trade-off across various predicates, we adopt the Top-Down strategy as it yields the highest Mean score.

\section{Conclusion and Future Work}\label{sec5}

In this paper, we revealed the drawbacks of relying solely on spatial features based on bboxes and discovered that the key to PSG task lies in the integration of shape-aware features. Thus, we proposed a model-agnostic CAFE framework that integrates shape-aware features in an easy-to-hard manner. Specifically, our approach deployed three classifiers, each specialized to handle predicates with increasing learning difficulties and equipped with corresponding sets of features of ascending complexity. Comprehensive experiments on the challenging PSG dataset showed that CAFE significantly improves the performance of both robust PSG and zero-shot PSG. In the future, we would like to extend CAFE to panoptic video scene graph generation task to construct comprehensive real-world visual perception systems.

\noindent
\\
\small
\noindent\textbf{Data Availability}
All experiments are conducted on publicly available datasets; see the references cited.

\bibliographystyle{spbasic}      
\bibliography{egbib.bib}


\newpage
\appendix

\section*{Appendix}
\appendix
\modified{This appendix is organized as follows:
\pagestyle{empty}
\thispagestyle{empty} 
\begin{itemize}
\item Detailed performance comparison analyses are presented in Sec.~\ref{sec:a}.
\item Statistics of computation cost and parameters are provided in Sec.~\ref{sec:b}.
\item Limitation is discussed in Sec.~\ref{sec:c}.
\end{itemize}}

\section{Detailed Performance Comparison Analyses}
\label{sec:a}
\subsection{Performance of Each Component of CAFE}

\begin{figure*}[t]
  \centering
  \includegraphics[width=0.85\linewidth]{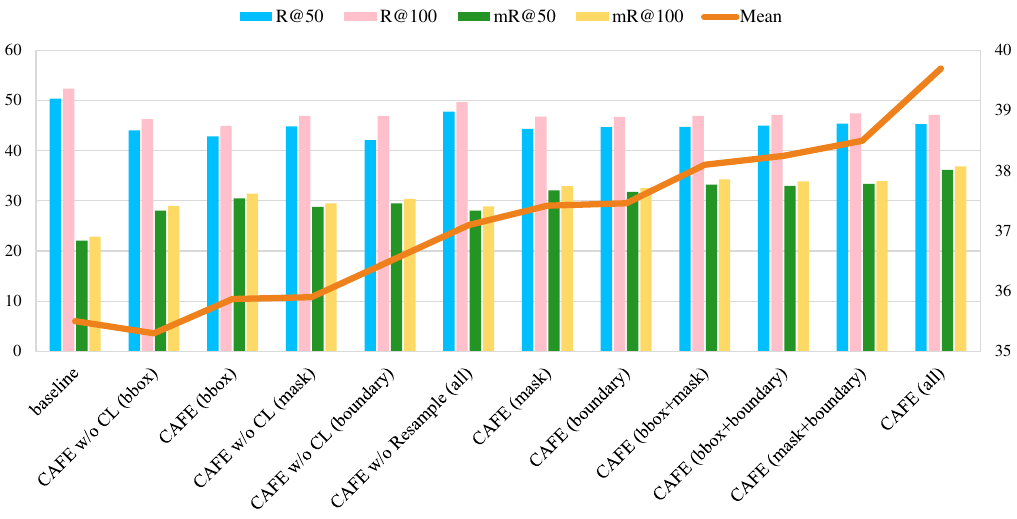}
  \caption{\modified{Performance comparison of each component of CAFE against the baseline. The baseline model is Motifs~\cite{zellers2018neural} under the PredCls setting.}} 
  \label{fig:performance}
\end{figure*}
\modified{To better illustrate the reasons behind performance improvements, we visualized the performance of each component within the CAFE model, under the Motifs~\cite{zellers2018neural} in the PredCls setting in Fig.~\ref{fig:performance}. The metrics used for performance comparison are R@50/100 and mR@50/100, while the \textcolor[RGB]{239,129,29}{orange} line represents the Mean metric across different configurations. 
For conciseness, we denote ``CAFE w/o CL" to indicate the model variant that solely employs the resampling strategy, and ``CAFE w/o Resample" for the variant that exclusively uses curriculum learning. The features utilized by each CAFE component are specified within the parentheses.

As shown in Fig.~\ref{fig:performance}, we can have the following observations: 
\begin{itemize}[leftmargin=*]
    \item[$\bullet$] \textbf{The significance of predicate resampling strategy.} In scenarios where shape-aware features are absent (\ie, only using bbox feature), the resampling strategy exhibits performance degradation on the head predicates (\ie, R@K), while yielding a slight improvement on the tail predicates (\ie, mR@K). This occurs because the resampling strategy aims to mitigate the extreme dataset imbalance by undersampling the head predicates and oversampling the tail predicates. The bbox feature's strength in identifying simple positional relations, primarily in head predicates, results in a marked decline in the R@K metric. However, the introduction of shape-aware features (\ie, mask and boundary features) turns the resampling strategy into a positive force for performance enhancement, achieving a higher Mean metric than the baseline. Crucially, when these three features are incorporated into the model training via a curriculum learning approach, the resampling strategy manifests a significant advantage. This is attributed to shape-aware features' ability to capture the shape of objects and the interaction information between object pairs, thereby effectively aiding in relation prediction. Moreover, balanced data can effectively prevent biased predictions and enhance the overall performance. We also conducted ablations on predicate sampling strategies in Table~\ref{ablation: resampling}.
    \item[$\bullet$]   \textbf{The importance of curricular feature training.} Curricular feature training constitutes a pivotal innovation in CAFE, playing a crucial role in enhancing model performance (\eg, higher Mean). Taking the use of a single feature as an example, the integration of the curriculum learning approach significantly improves the model's ability to predict tail predicates (\eg, higher mR@K). This improvement stems from the introduction of cognition-based predicate grouping and the division of the training process into three distinct stages. The former segregates predicates with semantic similarity into different groups, effectively mitigating semantic confusion within the relation classifier. The latter assigns a dedicated relation classifier to each stage and configures the classification space, ensuring that each classifier demonstrates strong discriminative abilities. Moreover, since our curriculum feature training incorporates knowledge distillation to preserve knowledge acquired in earlier stages, it can maintain the performance of head predicates (\eg, competitive R@K). We also performed detailed analyses through ablation studies on curriculum learning in Table~\ref{ablation: cl}.
    \item[$\bullet$]   \textbf{The combination of shape-aware features.} The integration of shape-aware features into PSG marks a notable advancement, as we not only pinpoint a crucial flaw in current PSG research (\ie, over-reliance on tight bounding boxes), but also offer a novel perspective on addressing segmentation challenges. Given that CAFE comprises a training process with three stages, each tailored to a specific set of features, the combination of these features is crucial for the model's performance. Firstly, the combination of any two features generally shows superior performance compared to using each feature on its own, which indicates a synergistic effect between the features. Secondly, incrementally adding complexity to the features leads to a gradual improvement in the Mean metric. Lastly, integrating all features culminates in achieving the best mR@K and the highest Mean score. This occurs because CAFE structures its training process in phases, progressively increasing the complexity of features in accordance with the rising difficulty of predicates. For instance, the boundary feature, capable of capturing interactions between object pairs, is introduced in the third stage to better predict relations of higher cognitive difficulty. Hence, by amalgamating all features, we capitalize on the distinct advantages of each to predict predicates accurately within their targeted groups, maximizing the performance to its fullest potential. Additionally, the ablation studies on different features can be found in Table~\ref{ablation:features}.
\end{itemize}
}

\subsection{Performance Comparison over All Predicates}
\begin{figure}[!t]
  \centering
  \includegraphics[width=\linewidth]{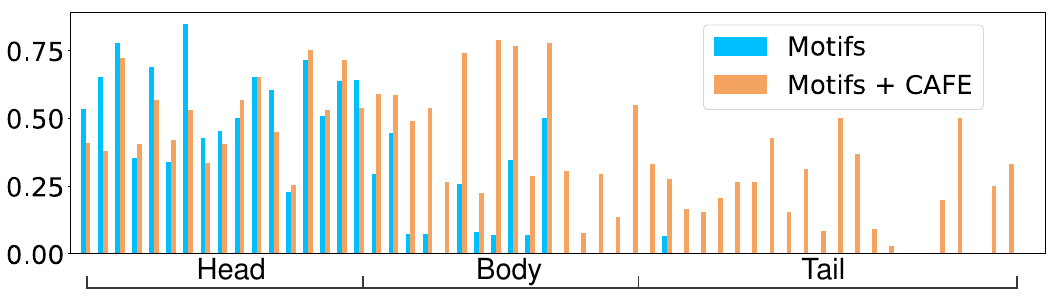}
  \caption{R@100 for each predicate category under the PredCls setting. The baseline model is Motifs~\cite{zellers2018neural}.}
  \label{fig:R@100_predcls}
\end{figure}
\begin{figure}[!t]
  \centering
  \includegraphics[width=\linewidth]{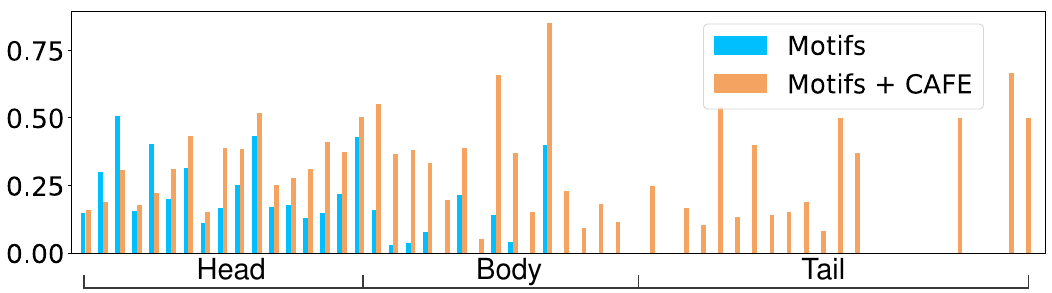}
  \caption{R@100 for each predicate category under the SGDet setting. The baseline model is Motifs~\cite{zellers2018neural}.}
  \label{fig:R@100_sgdet}
\end{figure}
To provide a more comprehensive illustration of model performance over all predicates, we presented recall statistics for each predicate using CAFE and Motifs~\cite{zellers2018neural} under both the PredCls and SGDet settings, as depicted in Fig.~\ref{fig:R@100_predcls} and Fig.~\ref{fig:R@100_sgdet}. We divided predicates into ``Head", ``Body", and ``Tail" groups based on the distribution in different ground-truth annotations in the PSG dataset. Notably, our proposed CAFE model demonstrates a favorable balance across various predicate categories in both task scenarios. To delve into specifics, while a slight decrease is observed in the head group, CAFE exhibits noteworthy enhancements in both the body and tail groups.

\subsection{Performance under Different Panoptic Segmentors}

\begin{table*}[!t]
\begin{center}
  \resizebox{.9\textwidth}{!}{
    \setlength{\tabcolsep}{1.0mm}
\begin{tabular}{lc||ccc|ccc|c|c}
\specialrule{0.05em}{0pt}{0pt}     \hline
\multicolumn{1}{c}{\multirow{2}{*}{Models}} & \multicolumn{1}{c||}{\multirow{2}{*}{Panoptic Segmentor}} & \multicolumn{8}{c}{SGDet} \\
\cline{3-10}    \multicolumn{2}{c||}{} & \multicolumn{1}{c}{R@20} &\multicolumn{1}{c}{R@50}&\multicolumn{1}{c|}{R@100} & \multicolumn{1}{c}{mR@20}& \multicolumn{1}{c}{mR@50} & \multicolumn{1}{c|}{mR@100} & \multicolumn{1}{c|}{Mean} & \multicolumn{1}{c}{PQ}\\
    \hline
    \hline
    VCTree~\cite{tang2019learning} & \multirow{3}{*}{Panoptic FPN~\cite{kirillov2019panoptic}}&
    20.6  & 22.1  & 22.5  & 9.7   & 10.2  & 10.2  & 15.9  & 40.2  \\
    \ \ +RCpsg~\cite{DBLP:conf/mm/YangWLWWYC23} & \multicolumn{1}{c||}{} &  22.3  & 24.2  & 24.6  & 11.0  & 11.5  & 11.8  & 17.6  & \textbf{40.7}  \\
     \ \ +CAFE (Ours) & \multicolumn{1}{c||}{} &  17.6  & 19.4  & 19.9  & \textbf{17.6}  & \textbf{18.1}  & \textbf{18.3}  & \textbf{18.5 } & 40.3  \\
    \hline
    VCTree~\cite{tang2019learning} & \multirow{3}{*}{Panoptic SegFormer~\cite{li2022panoptic}} &  26.4  & 28.7  & 29.4  & 12.3  & 13.2  & 13.4  & 20.6  & 49.4  \\
    \ \ +RCpsg~\cite{DBLP:conf/mm/YangWLWWYC23} & \multicolumn{1}{c||}{} & 29.1  & 30.9  & 31.4  & 13.8  & 14.5  & 14.7  & 22.4  & 49.7  \\
    \ \ +CAFE (Ours) & \multicolumn{1}{c||}{} &   26.9  & 29.8  & 30.6  & \textbf{27.8}  & \textbf{29.1}  & \textbf{29.4}  & \textbf{28.9}  & \textbf{54.9}    \\
    \specialrule{0.05em}{0pt}{0pt}     \hline
    \end{tabular}
    }
\end{center}
\caption{\modified{Comparison of different models on the PSG dataset using different panoptic segmentors with the ResNet-50 backbone. The baseline model is VCTree~\cite{tang2019learning} under the SGDet setting.}}
\label{tab:segmentor}%
\end{table*}%

\begin{table*}[!t]
\begin{center}
  \resizebox{.9\textwidth}{!}{
    \setlength{\tabcolsep}{0.7mm}
\begin{tabular}{ll||ccc|ccc|c|c}
\specialrule{0.05em}{0pt}{0pt}     \hline
\multicolumn{1}{c}{\multirow{2}{*}{Models}} & \multicolumn{1}{c||}{\multirow{2}{*}{Panoptic Segmentor}} & \multicolumn{8}{c}{SGDet} \\
\cline{3-10}    \multicolumn{2}{c||}{} & \multicolumn{1}{c}{R@20} &\multicolumn{1}{c}{R@50}&\multicolumn{1}{c|}{R@100} & \multicolumn{1}{c}{mR@20}& \multicolumn{1}{c}{mR@50} & \multicolumn{1}{c|}{mR@100} & \multicolumn{1}{c|}{Mean} & \multicolumn{1}{c}{PQ}\\
    \hline
    \hline
 \multirow{2}{*}{Motifs+CAFE} & Panoptic SegFormer~\cite{li2022panoptic} & 23.4 & 25.7 & 26.4 & 25.5 & 26.5 & 26.8 & 25.7  & 54.9\\
         & Mask2Former~\cite{DBLP:conf/cvpr/ChengMSKG22} &   24.8  & 27.2  & 27.8  & 28.1  & 29.2  & 29.6  & 27.8  & 55.4\\
    \hline
    \multirow{2}{*}{Transfomrer+CAFE} & Panoptic SegFormer~\cite{li2022panoptic} & 24.6 & 27.6 & 28.7 & 25.0 & 26.6 & 26.9 & 26.6 & 54.9\\
         & Mask2Former~\cite{DBLP:conf/cvpr/ChengMSKG22} & 26.0  & 28.0  & 28.6  & 28.1  & 29.0  & 29.2  & 28.1   & 55.4\\
    \specialrule{0.05em}{0pt}{0pt}     \hline
    \end{tabular}
    }
\end{center}
\caption{\modified{Performance (\%) of CAFE on the PSG dataset with different panoptic segmentors under the SGDet setting across two baselines.}}
\label{tab:detector}%
\end{table*}%
\modified{To demonstrate the influence of panoptic segmentors with varying capabilities in generating panoptic segmentation masks, we adopted Panoptic FPN~\cite{kirillov2019panoptic} and Panoptic Segformer~\cite{li2022panoptic} as panoptic segmentors. Subsequently, we compared performance of CAFE with VCTree~\cite{tang2019learning} and RCpsg~\cite{DBLP:conf/mm/YangWLWWYC23} under the SGDet setting. As shown in Table~\ref{tab:segmentor}, we can observe that: 1) The performance of all three models is influenced by the accuracy of the panoptic segmentation masks. When the quality of masks improves, the overall performance of the models also improves. 2) Even when the panoptic segmentation masks are of low quality (\eg, Panoptic FPN), CAFE consistently produces superior performance compared to the baseline models, which indicates that CAFE has a relatively low dependency on panoptic segmentation masks. This is because the performance benefits of CAFE are not solely derived from shape-aware features but are also influenced by the resampling strategy and curricular training mode. 3) With the improvement in mask accuracy, CAFE demonstrates larger performance gains. This is attributed to the fact that when the quality of panoptic segmentation masks improves, the extracted shape-aware features achieve higher precision. This enhancement enables the model to more effectively capture interactions between objects, thereby resulting in improved performance.

\begin{table*}[t]
  \begin{center}
  \resizebox{0.9\textwidth}{!}{
    \setlength{\tabcolsep}{1.2mm}
    \begin{tabular}{l||c|c|c|c|c|c}
    \specialrule{0.05em}{0pt}{0pt}
    \hline
    \multicolumn{1}{c||}{\multirow{2}{*}{Models}} & \multicolumn{3}{c|}{PredCls} & \multicolumn{3}{c}{SGDet}\\
\cline{2-7}          & Training Cost (s)  & Test Cost (s) & Params (M)  & Training Cost (s) & Test Cost (s)  & Params (M) \\
    \hline
    \hline
    Motifs~\cite{zellers2018neural} & 1.28 & 0.089 & 104.0 &  2.08 & 0.245 & 103.7\\
    \ \ +CAFE   & 1.68  & 0.137 & 167.8 &  2.34 & 0.256 & 167.6   \\
    \hline
    VCTree~\cite{tang2019learning} & 2.71  & 0.167 & 99.7 &  4.04 & 0.236 & 99.5\\
    \ \ +CAFE   & 3.24  & 0.303 & 154.2 & 5.30  & 0.213 & 154.9   \\
    \hline
    Transformer~\cite{vaswani2017attention} & 1.91  & 0.082 & 100.2 & 2.15  & 0.119 & 100.2\\
    \ \ +CAFE & 2.38  &  0.204 & 139.7  & 3.55 & 0.165 &  140.4  \\
    \specialrule{0.05em}{0pt}{0pt}
    \hline
    \end{tabular}
    }
    \end{center}
    \caption{\modified{The Statistics of training cost, test cost and parameters of two-stage methods, under both PredCls and SGDet settings.}}
  \label{tab:cost}
\end{table*}%

Additionally, we also selected Mask2Former~\cite{DBLP:conf/cvpr/ChengMSKG22}, which can generate higher quality masks, as one of the panoptic segmentors. We conducted experiments under the SGDet setting across two baselines (\ie, Motifs~\cite{zellers2018neural} and Transformer~\cite{vaswani2017attention}). From the results shown in Table~\ref{tab:detector}, it is evident that when the panoptic segmentor generates panoptic segmentation masks with higher quality or accuracy (\ie, higher PQ), CAFE exhibits a substantial improvement in model performance (\eg, 26.8\% using Panoptic SegFormer \vs 29.6\% using Mask2Former in mR@100 based on Motifs). As computer vision technology continues to evolve, we believe that more accurate panoptic segmentation models will emerge, enabling CAFE to have a greater impact on PSG tasks. In the future, we will explore the combination of CAFE with the SAM~\cite{DBLP:conf/iccv/KirillovMRMRGXW23} model, thereby enhancing the robustness and generalization capabilities of CAFE.}

\section{Statistics of Computation Cost and Parameters}
\label{sec:b}
\modified{To compare CAFE with other PSG methods in terms of computational cost and parameters, we tested various two-stage models for the expenses (s) during both the training and inference phases per image, as well as the training parameters (M) in Table~\ref{tab:cost}. From the results in Table~\ref{tab:cost}, we can observe that: 1) Compared to two-stage baseline models, although our proposed CAFE exhibits a slight increase in training and test costs, these minor increases are justified by the substantial performance improvement. 2) Although our CAFE also experiences a slight increase in model parameters, all experiments can be completed on a single NVIDIA 2080Ti GPU, with training time for one epoch being less than 5 hours. We also visualized the GPU consumption and total training cost among different PSG models in Fig.~\ref{fig:cost}.}

\section{Limitations}
\label{sec:c}
Although CAFE is a model-agnostic approach seamlessly integrable into any advanced PSG architecture, it is essential to clarify that in this context, ``PSG architecture" refers specifically to all two-stage frameworks. This distinction arises from our current methodology of extracting shape-aware features, which derives from masks generated by the panoptic segmentor. 
Moreover, currently, we are constrained by insufficient computational resources to facilitate the training of one-stage methods (\eg, PSGTR~\cite{yang2022panoptic}), demanding the utilization of eight V100 GPUs with a batch size of 1.  Given these considerations, it is not straightforward to apply CAFE to one-stage methods.

Besides, in order to retain the knowledge acquired in earlier phases, we incorporate knowledge distillation (\cf, Sec.~\ref{3.3}). It's worth noting that while knowledge distillation may introduce extra computational overhead or increased GPU consumption during training, it does not incur any additional overhead during the test stages. More importantly, the benefits of knowledge distillation are considered significant enough to justify the additional cost.

\end{sloppypar}
\end{document}